\pdfoutput=1

\documentclass[11pt]{article}
\usepackage{colortbl}
\usepackage{enumitem}
\usepackage{amsmath}
\usepackage[utf8]{inputenc}

\usepackage[final]{acl}

\usepackage{times}
\usepackage{latexsym}

\usepackage[T1]{fontenc}

\usepackage[utf8]{inputenc}

\usepackage{microtype}

\usepackage{inconsolata}

\usepackage{graphicx}

\usepackage{listings}

\title{I Learn Better If You Speak My Language: Understanding the Superior Performance of Fine-Tuning Large Language Models with LLM-Generated Responses}

\author{Xuan Ren\textsuperscript{\rm 1}, Biao Wu\textsuperscript{\rm 2}, Lingqiao Liu\textsuperscript{\rm 1}\thanks{\ \ Corresponding author.} \\
\textsuperscript{\rm 1}University of Adelaide,
\textsuperscript{\rm 2}University of Technology Sydney\\
xuan.ren@adelaide.edu.au, biaowu165534@gmail.com, lingqiao.liu@adelaide.edu.au
}

\date{}

\begin{document}
\maketitle
\begin{abstract}
This paper explores an intriguing observation: fine-tuning a large language model (LLM) with responses generated by a LLM often yields better results than using responses generated by humans, particularly in reasoning tasks. We conduct an in-depth investigation to understand why this occurs. Contrary to the common belief that these instances is due to the more detailed nature of LLM-generated content, our study identifies another contributing factor: an LLM is inherently more "familiar" with LLM generated responses. This familiarity is evidenced by lower perplexity before fine-tuning. We design a series of experiments to understand the impact of the "familiarity" and our conclusion reveals that this "familiarity" significantly impacts learning performance. Training with LLM-generated responses not only enhances performance but also helps maintain the model's capabilities in other reasoning tasks after fine-tuning on a specific task. Our code and data are public at \href{https://github.com/XuanRen4470/I-Learn-Better-If-You-Speak-My-Language}{https://github.com/XuanRen4470/I-Learn-Better-If-You-Speak-My-Language}.

\end{abstract}

\section{Introduction}

Recent research has demonstrated that a large language model (LLM) can generate training data for another LLM \cite{dai2023auggpt, edwards-etal-2022-guiding, moller-etal-2024-parrot, guo2023improving, ubani2023zeroshotdataaug, piedboeuf-langlais-2023-chatgpt, agrawal2023qameleon,xu2023wizardlm,kieser2023educational, alpaca, peng2023instruction,xu2024wizardlm}. This approach offers a method for transferring knowledge from a larger model to a smaller one, or for creating supplementary training materials, such as rationales \cite{zhang2024distillation,kang2023knowledgeaugmented,li2022explanations}, or refined step-by-step reasoning \cite{hsieh2023distilling,ho2022large,magister-etal-2023-teaching,fu2023specializing}.

\begin{figure*}
  \centering
  \includegraphics[width=1.0\textwidth]{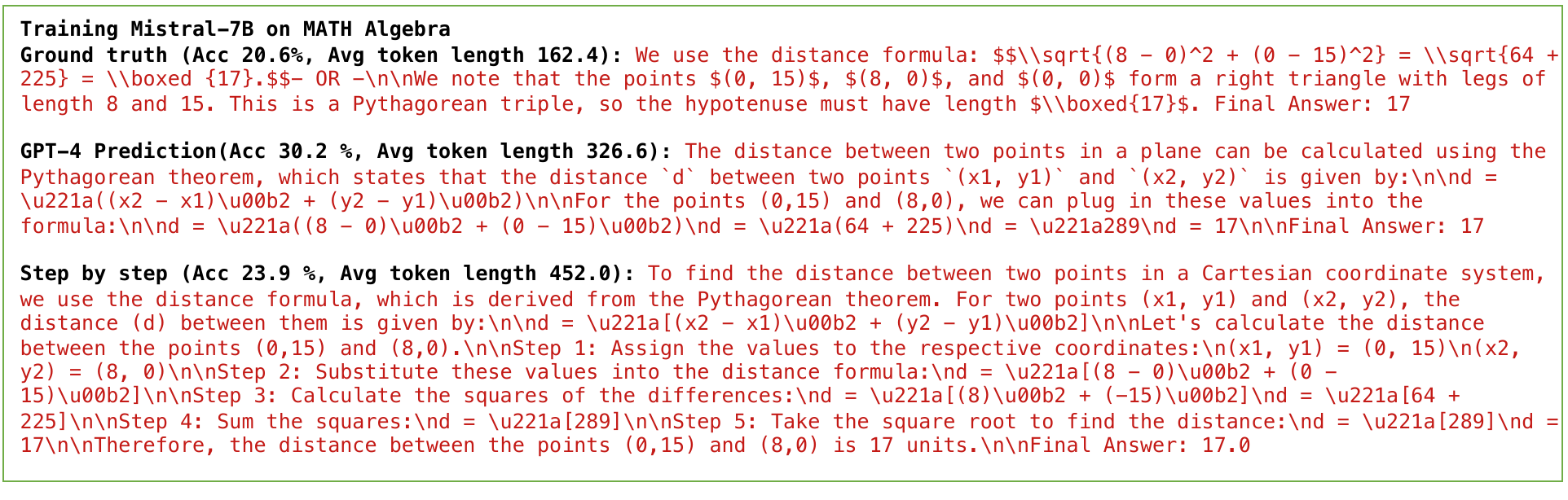}
  \label{fig:figure1}
\caption{This figure shows training outcomes for different data generation methods, demonstrating that more details do not always improve performance.}
\end{figure*}

In this paper, we made an intriguing observation that fine-tuning an LLM (dubbed target LLM) with responses generated by a LLM, either target LLM or other LLMs, often yields better results than using responses provided by humans in chain-of-thought reasoning tasks, as shown in \hyperref[tab:table1]{Table 1}. The prevailing view attributes the successful training results of LLM-generated data to the fact that LLMs provide more details than human-annotated data, such as `chain-of-thought-style' reasoning process\cite{hsieh2023distilling,ho2022large,magister-etal-2023-teaching,fu2023specializing}.
However, our experiments have found numerous counterexamples. As shown in \hyperref[fig:figure1]{Figure 1}
, target responses that include more details and utilize chain of thought styles do not necessarily result in better training outcomes. This indicates there are other factors contributing to the excellent performance of LLM generated responses.



The observation above motivates us to investigate the underlying reason for this phenomenon. We found that whether the target LLM is familiar with the target responses is a crucial factor influencing model training on reasoning tasks. 

Our main hypothesis are as follows:

\begin{itemize}[noitemsep, topsep=0pt, partopsep=0pt]
    \item Given the same question, the target LLM is inherently familiar with responses generated by itself or other LLMs.
    \item The target LLM can better adapt to and leverage responses in its familiar styles, leading to improved outcomes.
\end{itemize}

The following section of this paper will examine the hypothesis through a series of experiments. We will measure `familiarity,' conduct ablation studies by creating different variants of LLM-generated responses, and evaluate their performance to understand various factors in the process. This will provide insights into our hypothesis. 

\section{Empirical Study Protocols}
The purpose of this paper is to understand and investigate the aforementioned phenomenon 
through empirical analysis. We begin by introducing our study protocols, including the datasets and evaluation approaches used.

\subsection{Training and Evaluation approaches}
Without loss of generality, we denote the training data as question and response pairs $(q,r)$. We will use a subset of the Llama2 benchmark \cite{touvron2023llama} to create the training dataset. We will use a subset of the Llama2 benchmark that contains human-labeled explicit reasoning processes to create the training dataset. The question and response pairs in these datasets are human-constructed. Additionally, we will use an LLM to generate new responses, following $r'=LLM(q,P)$, where $LLM$ denotes the language model and $P$ denotes the prompt used to create the response. We will explore different $P$ in the following experiment. This creates a synthetic version of dataset $\{(q,r')\}$.

Our training approach for the Llama2-13b-chat\cite{touvron2023llama} and Mistral-7b-Instruct-v2\cite{jiang2023mistral} models involves specific adaptations to optimize performance. For general, fully supervised fine-tuning of the Llama2-13b-chat and Mistral-7b-Instruct-v2 models, we employ a default learning rate of 2e-5, a batch size of 32, and a warm-up period comprising 10\% of the total steps. In contrast, Given the small size of the training datasets, such as MBPP and HumanEval, we reduced the batch size to 10 to accommodate the limited data. For the Llama2-13b-chat model training on the HumanEval dataset, we set the learning rate to 2e-4. This adjustment was necessary as lower rates proved insufficient for the model to effectively learn from this dataset. Additionally, we use a cosine learning rate schedule and fine-tune only the Q and V matrices of the LORA\cite{hu2021lora} parameters with a rank of 8. All models are trained and evaluated using half-precision.

We evaluate the fine-tuned model in two different ways. One is called in-domain evaluation, which assesses performance on the tasks of the training set. The other is called cross-task evaluation, which evaluates performance across tasks on other datasets. The purpose of using cross-task evaluation is to test if the model can benefit from the training process of a similar task and if the model will forget its capability on other tasks after training on a specific task.

We have developed an evaluation script to automatically measure and report the pass@1 accuracy of models for each dataset. For the HumanEval dataset, we utilize its built-in testing script to ensure the accuracy of results. Our evaluation scripts require answers to follow a specific format, detailed in Appendix \ref{appendix:evalscript}. The models not trained in this format have demonstrated difficulty adhering to it for coding tasks, such as MBPP and HumanEval \cite{chen2021evaluating, austin2021program}. Therefore, cross-domain evaluations are not conducted for MBPP and HumanEval.

\subsection{Dataset and API}


We utilized the Llama2 benchmark \cite{touvron2023llama} to select five datasets that include a detailed reasoning process in their annotations. This category includes GSM8K\cite{cobbe2021training}, MATH\cite{hendrycks2021measuring}, HumanEval\cite{chen2021evaluating}, MBPP\cite{austin2021program}, and ECQA \cite{aggarwal2021explanations}, where the ground truth comprises a complete human-labeled problem-solving process or rationale, not just the final answer. For example, the ground truth in GSM8K involves comprehensive mathematical reasoning, and the coding datasets’ ground truths consist of complete codes, which are logical sequences by nature. We included coding datasets like MBPP and HumanEval in this category, as developing code inherently involves logical reasoning, even though it may not always be presented in the form of a traditional chain of thought. Our model was trained on one dataset and subjected to in-domain testing on the same dataset, as well as cross-domain testing on other datasets. Our selection includes all coding and mathematics datasets within the Llama2 benchmark. We chose to use the ECQA dataset because it includes human annotated rationales, unlike the CQA\cite{talmor2018commonsenseqa} dataset included in the Llama2 benchmark.

\textbf{Code}: We evaluated coding ability using the HumanEval \cite{chen2021evaluating} and MBPP \cite{austin2021program} datasets. Note that the HumanEval dataset contains only 164 testing data points with no training data. We divided the training task into two parts: in the first part, we trained on the first 82 data points and tested on the last 82. In the second part, we reversed this order. Ultimately, we averaged the test results from both parts. For the MBPP-full dataset, as opposed to the MBPP-sanitized version which includes fewer training data points, we utilized all available training data (374 data points) and testing data (500 data points)

\textbf{Commonsense Reasoning}: For Commonsense QA task, we selected the ECQA \cite{aggarwal2021explanations} dataset, which provides explanations for the answers from CQA\cite{talmor2018commonsenseqa} dataset. We combined these explanations with the multiple-choice answers to form the ground truth training data. We used the first 1000 training and test data points for training and evaluation.

\textbf{Math}: We conducted training and evaluations on the GSM8K \cite{cobbe2021training} and MATH \cite{hendrycks2021measuring} datasets. The MATH dataset encompasses several categories. We conducted training and evaluations exclusively within the algebra category. For the evaluation, we developed a script that compares the ground-truth answer with the model's final prediction. This script accurately evaluates responses that are numerical. However, in the MATH dataset, some answers contain inequalities or complex mathematical expressions. We filtered out these data points and used only those with numerical answers for both training and evaluation. After excluding data points with non-numerical answers, we used the first 1000 training data points for training and all the testing data points for testing. As a result, we obtained 1000 training and 1314 testing data points for GSM8K, and 1000 training and 752 testing data points for MATH (algebra).

\textbf{API}: We did our experiment with use gpt-4-1106-preview \cite{openai_gpt4_api} from OpenAI and claude-3-5-sonnet-20240620 \cite{anthropic_claude_api} from Anthropic.

\section{Observation and Hypothesis}

\subsection{Observation 1: Superior Performance of LLMs Over Human-Annotated Labels}

Our study is inspired by a compelling observation: when an advanced LLM, such as GPT-4 or Claude 3.5, is used to generate responses to questions, and these questions along with their generated responses are used as a synthetic dataset, fine-tuning a smaller LLM (such as Llama2-13B-Chat or Mistral-7B-Instruct-v2) on this synthetic dataset often yields better performance than using the original dataset with human-provided answers. The detailed experimental results are presented in \hyperref[tab:table1]{Table 1}. As demonstrated, the majority of the low-performance data points (14 out of 17), marked in red, reflect accuracies when training with human-annotated ground truth. Using synthetic datasets consistently yields higher in-domain performance compared to human-answered datasets. Notably, we observe a significant improvement in math-related tasks, often exceeding a 10\% absolute increase in in-domain performance. Additionally, the model trained on the synthetic data achieves higher cross-domain performance. This observation motivates us to investigate the underlying reasons.


\subsection{Observation 2: Significantly Lower Perplexity of LLM generated responses Over Human-Annotated Labels}

\begin{figure*}
  \centering
  \includegraphics[width=1\textwidth]{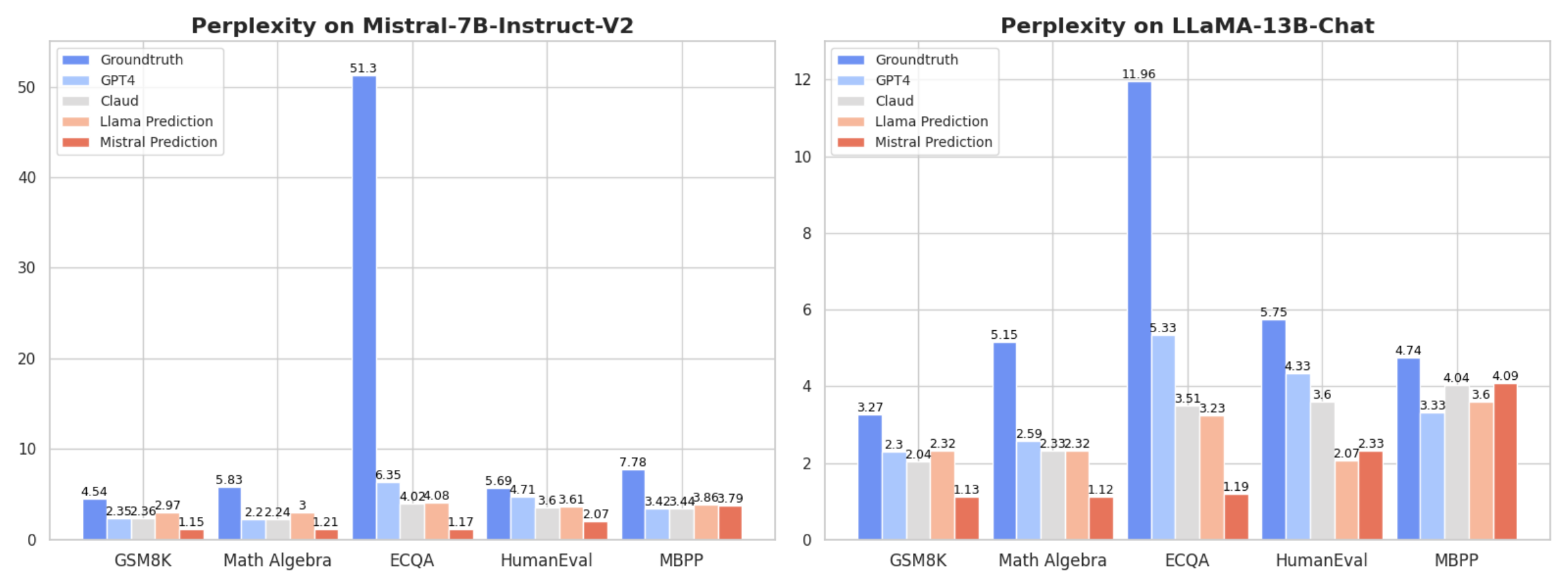}
  \caption{Average Perplexity Comparison}
  \label{fig:figure2}
\end{figure*}

We measured the perplexity of responses as assessed by the target LLMs (Mistral-7B-Instruct-v2 and Llama2-13B-chat). An interesting observation, shown in \hyperref[fig:figure2]{Figure 2}, is that responses generated by LLMs such as GPT-4, Claude 3.5, Mistral-7B-Instruct-v2, and Llama2-13B-chat, using a zero-shot approach, exhibit significantly lower perplexity compared to those provided by humans for the same input questions from the training data. Lower perplexity suggests that LLMs are more familiar with responses generated by other LLMs or themselves than with human-generated responses.

\subsection{Hypotheses}
From Observation 2, we notice that given the same question, the average perplexity of responses generated by LLMs is consistently lower than the perplexity of ground truth annotated by humans. This observation leads us to pose our first hypothesis: When generating target responses for the same question, one LLM inherently more familiar with data generated by another LLM. We not only observed consistently lower perplexity from advanced LLMs (GPT-4 and Claude 3.5) across tasks in Observation 2 but also noted that LLMs trained on advanced LLMs generated responses consistently outperform the same LLMs trained using human-annotated data. This phenomenon leads us to propose our second hypothesis: An LLM performs better when trained on data it is familiar with.

\begin{table*}[ht]
\centering
\resizebox{\textwidth}{!}{
\begin{tabular}{l|l|c|c|c|c|c|c|c|}
\hline

Method & Training Dataset and Model Type& GSM8K & Math Algebra  & ECQA &HumanEval & MBPP   \\ \hline

Groundtruth & GSM8K, Mistral&    \cellcolor{gray!50} \textcolor{red}{0.434}& \textcolor{red}{0.162}&0.594  && \\
GPT-4 Answer Directly & &               \cellcolor{gray!50} 0.597  &0.246  & 0.597   &&\\
Claude Answer Directly & &               \cellcolor{gray!50}  0.586 & 0.230  &0.595 & & \\

\hline

Groundtruth & GSM8K, Llama&    \cellcolor{gray!50}\textcolor{red}{0.364} & 0.141& 0.565 &&\\
GPT-4 Answer Directly & &               \cellcolor{gray!50} 0.428  &0.128  &  0.575 && \\
Claude Answer Directly &  &               \cellcolor{gray!50}  0.433 &  \textcolor{red}{0.110} & 0.548 & & \\

\hline

Groundtruth & Math algebra, Mistral&   \textcolor{red}{0.264}  &\cellcolor{gray!50}\textcolor{red}{0.206} & 
0.554  && \\
GPT-4 Answer Directly  & &         0.553    & \cellcolor{gray!50} 0.302&0.608  &&   \\
Claude Answer Directly & &      0.554      & \cellcolor{gray!50}0.277  &0.606 & & \\

\hline

Groundtruth & Math algebra, Llama&  0.36   & \cellcolor{gray!50}\textcolor{red}{0.126}&0.575&& \\
GPT-4 Answer Directly  & &     0.35    & \cellcolor{gray!50}0.150& 0.561 &&  \\
Claude Answer Directly & &   0.317     & \cellcolor{gray!50} 0.137 &0.54 & & \\

\hline

Groundtruth & ECQA, Mistral&  \textcolor{red}{0.258} &\textcolor{red}{0.134} &   \cellcolor{gray!50}0.68  &&  \\
GPT-4 Answer Directly & &               0.462  &  0.223&     \cellcolor{gray!50}0.722&&\\
Claude Answer Directly & &     0.457    &0.213  & \cellcolor{gray!50} 0.714 & & \\

\hline

Groundtruth & ECQA, Llama&   \textcolor{red}{0.132}  & \textcolor{red}{0.0798}&  \cellcolor{gray!50}\textcolor{red}{0.631} && \\
GPT-4 Answer Directly & &             0.379   &  0.156&    \cellcolor{gray!50}0.656 &&  \\
Claude Answer Directly & &    0.38   & 0.129 & \cellcolor{gray!50} 0.678 & & \\

\hline

Groundtruth & HumanEval, Mistral& 0.363    & 0.191&0.583   &  \cellcolor{gray!50}0.323 &   \\
GPT-4 Answer Directly &            & 0.313 &  0.163 &   0.581  &\cellcolor{gray!50}0.366    &   \\
Claude Answer Directly & &  0.383     & 0.181 & 0.553&  \cellcolor{gray!50} 0.372 & \\

\hline

Groundtruth & HumanEval, Llama&  \textcolor{red}{0.0705} &	\textcolor{red}{0.083} & 0.528&    \cellcolor{gray!50} 0.146 &  \\
GPT-4 Answer Directly & &  \textcolor{red}{0.125} & 0.105 &  0.552 &  \cellcolor{gray!50} 0.159&   \\
Claude Answer Directly & &     0.296  &0.104  &   \textcolor{red}{0.435} &  \cellcolor{gray!50} 0.140& \\

\hline

Groundtruth & MBPP, Mistral&      0.392& 0.176&  0.527   & &\cellcolor{gray!50} \textcolor{red}{0.276} \\
GPT-4 Answer Directly & &        0.399       & 0.186 & 0.568   &    & \cellcolor{gray!50} 0.354 \\
Claude Answer Directly & &     0.439   & 0.189 &0.573 &  & \cellcolor{gray!50} 0.37 \\

\hline

Groundtruth & MBPP, Llama&   0.328 & 0.132& 0.556    & &\cellcolor{gray!50}  0.2  \\
GPT-4 Answer Directly & &  0.351      & 0.140 &  0.574    &   & \cellcolor{gray!50} 0.202  \\
Claude Answer Directly & &             0.339   & 0.122 &0.546 & &  \cellcolor{gray!50} 0.204\\

\hline

\end{tabular}
}

\label{tab:table1}
\caption{Human-annotated data Vs. data generated directly by GPT-4/Claud 3.5. In-domain performance is highlighted in grey. Data points are highlighted when accuracy is more than 15\% below the highest accuracy on the same dataset and model. There are 14, 1, and 2 red data points for Groundtruth, GPT-4, and Claude, respectively.}
\end{table*}

\section{Investigation of the hypotheses}

\begin{table*}[h]
\centering
\resizebox{\textwidth}{!}{
\begin{tabular}{l|l|c|c|c|c|c|c|c}
\hline

Method & Training Dataset and Model Type& GSM8K & Math Algebra  & ECQA & token length  \\ \hline

Zeroshot Mistral & &    0.413 & 0.185&0.504   &\\
Zeroshot Llama2 & &    0.35 & 0.125&0.597  &\\
 \hline

GPT-4 Answer Directly  &Mistral &               \cellcolor{gray!50} 0.597  &0.246  & 0.597    &179.94\\

GPT-4 Step-by-step &            &     \cellcolor{gray!50} 0.574   & 0.224 & 0.591 &282.9\\
GPT-4 Detailed Step-by-Step transformation of GT &           &  \cellcolor{gray!50}  \textcolor{red}{0.506}      &  \textcolor{red}{0.196} & 0.602   &  315.028\\

\hline

GPT-4 Answer Directly &Llama &               \cellcolor{gray!50} 0.428  &0.128  &  0.575  &179.941\\

GPT-4 Step-by-step &            &     \cellcolor{gray!50}  0.402  & 0.134 &  0.553   &290.92\\
GPT-4 Detailed Step-by-Step transformation of GT   &           &  \cellcolor{gray!50}    0.396    & 0.114  & 0.553    &315.028\\ 

\hline

GPT-4 Answer Directly & Mistral&         0.553    & \cellcolor{gray!50} 0.302&0.608  & 328.76\\

GPT-4 Step-by-step &            &  0.533 &  \cellcolor{gray!50}  \textcolor{red}{  0.239}  & 0.583   &451.99\\
GPT-4 Detailed Step-by-Step transformation of GT &           &    \textcolor{red}{0.468}     & \cellcolor{gray!50} \textcolor{red}{0.223}  &0.582   &484.368 \\

\hline

GPT-4 Answer Directly& Llama&     0.35    & \cellcolor{gray!50}0.150& 0.561 & 328.76\\

GPT-4 Step-by-step &            &  0.344 &  \cellcolor{gray!50}  0.145   &  0.593    &450.08\\
GPT-4 Detailed Step-by-Step transformation of GT   &           &      \textcolor{red}{0.299}     & \cellcolor{gray!50}  \textcolor{red}{0.110} &0.545   &484.368\\ 

\hline

GPT-4 Answer Directly &Mistral &               0.462  &  0.223&     \cellcolor{gray!50}0.722&176.54\\

GPT-4 Step-by-step  &            & 0.481  &0.203 & \cellcolor{gray!50}  0.71      &325.15 \\
GPT-4 Detailed Step-by-Step transformation of GT  &           &    0.487    &   
\textcolor{red}{0.186} &   \cellcolor{gray!50}0.68 &  322.855\\

\hline

GPT-4 Answer Directly  &Llama &             0.379   &  0.156&    \cellcolor{gray!50}0.656 & 176.54\\


GPT-4 Step-by-step &            &  0.363 & 0.140& \cellcolor{gray!50} 0.648        & 337.30\\

GPT-4 Detailed Step-by-Step transformation of GT  &           &    \textcolor{red}{0.135}    &  \textcolor{red}{0.106}  &   \cellcolor{gray!50}0.66&322.855\\ 

\hline
\end{tabular}
}
\label{tab:table2}
\caption{Performance comparison of models trained on data constructed using different methods. $n_{\text{train}} = 1000$. Data points are labeled as low performance when accuracy is more than 15\% below the highest accuracy on the same dataset using the same model. See \hyperref[tab:table6]{Table 6} and \hyperref[tab:table7]{Table 7} for additional experiments with GPT-4 and Claude 3.5.}

\end{table*}


In this section, we conduct an in-depth investigation about the aforementioned hypotheses. Our investigations are organized into two parts. In Section \ref{sect: part1}, we examine to what extend the familiarity contribute to higher performance of using LLM generated responses. Particularly, we conduct experiments to ablate the factor that LLM could introduce additional reasoning details to the original human-provided response. In Section \ref{sec: style-transfer}, we experiment to determine if a model, when used alone without the assistance of a more advanced model, can rewrite ground truth into a form it is more familiar with, thereby enhancing training outcomes. The advantage of this experimental design is that it allows us to evaluate the effect of familiarity while removing any benefits that might arise from using a more advanced language model.

\subsection{Understand the impact of ``familiarity'' }\label{sect: part1}
\noindent \textbf{Experiment 1: Is observation 1 only rooted from more detailed reasoning steps provided by the LLM other than familiarity?}\\
Our observation 1 suggests that using advanced LLMs generated data leads to better training effectiveness than using human-annotated ground truth. Our second hypothesis posits that this enhanced learning by LLMs on LLMs responses could be attributed to familiarity with the data. It is a common belief that LLMs often includes more detailed content than human annotators, which could contribute to improved training outcomes. However, we question whether the superior results from advanced LLMs are solely due to the addition of more details, or if familiarity plays a significant role as well.

To investigate whether the superior performance is due to the additional detail provided by advanced LLMs compared to humans, we designed the following experiment. We employed different methods to have GPT-4 generate data in a detailed, step-by-step manner. For each dataset, we established a control group: one group involved GPT-4 transforming ground truth into its own detailed style, while the other had GPT-4 mimic the style of the human-labeled ground truth, adding detail to build upon it in a step-by-step style. As shown in \hyperref[tab:table2]{Table 2}, in the vast majority of cases, the training effectiveness of the first group, which used GPT-4's own style, was superior to that of the second group, which followed the human-annotated style. In addition, it is noteworthy that using the Mistral 7B model for training on math algebra, converting ground truth into step-by-step answers was even detrimental. Both of the step by step groundtruth transformation methods achieves accuracy below 24\% on Math Algebra, much lower than 30.2\%  achieved by `GPT-4 answer directly'(Gpt-4 answer directly represents employing GPT-4 to generate responses using only the questions from the training datasets. For the ECQA task, we provide GPT-4 with gold labels (excluding rationales) to guide its response generation.) We speculate that one possible reason for this is that adding `more than enough details' may sometimes complicate the model's reasoning process. We have included additional experiments in the appendix \ref{sect:gpt4_varient_appendix}, involving various GPT-4 (refer to \hyperref[tab:table6]{Table 6}) and Claude 3.5 (refer to \hyperref[tab:table7]{Table 7}) generated variants with different token lengths (shorter or longer), which generally conform to the patterns described above. For clarity and conciseness, \hyperref[tab:table2]{Table 2} only includes 2 examples of the step-by-step variants.

After preliminary validation of our hypotheses, we made another intriguing observation: According to \hyperref[tab:table2]{Table 2}, the `GPT-4 answer directly' dataset, characterized by a shorter average token length of target responses compared to both `step by step' and `step by step transformation of GT' datasets, suggests fewer details in its content. Despite this, the performance of the data directly generated by GPT-4 frequently ranks among the best of all GPT-4 answer variants. Meanwhile, other data generation methods exhibited significantly lower performance than GPT-4's direct answers on at least two tasks, as highlighted in red in \hyperref[tab:table2]{Table 2}. This suggests that, given a problem, the data directly generated by an advanced LLM might yield the good training results. We observed the same pattern in \hyperref[tab:table6]{Table 6} and \hyperref[tab:table7]{Table 7}. Artificially defined prompts that add information to the ground truth may lead to the generation of target responses that do not conform to the LLM’s linguistic or logical preferences.

\medskip
\noindent \textbf{Experiment 2: The impact of familiarity}

\begin{table*}[h]
\centering
\resizebox{\textwidth}{!}{
\begin{tabular}{l|l|c|c|c|c|c|c}
\hline

Method & Model Type& GSM8K& GSM8K Perplexity & Math Algebra & Math Algebra Perpleity & ECQA & ECQA Perplexity    \\ \hline

GPT-4 Answer with Lower Perplexity  &   Mistral         & 0.600   &2.20 &  0.303& 2.10 &0.722& 6.15   \\ 

GPT-4 Answer with Higher Perplexity &           &  0.547  & 5.58   &  \textcolor{red}{0.231}&4.66& 0.702 & 11.13 \\ 

\hline

GPT-4 Answer with Lower Perplexity &   Llama2         & 0.424  &  2.21 &  0.153 &  2.36&0.655&5.11   \\ 

GPT-4 Answer with Higher Perplexity&           &   0.38  &4.19  &  0.131 & 4.35&0.654 &7.69   \\ 

\hline

Claude Answer with Lower Perplexity  &   Mistral         & 0.586   &2.36    & 0.277  &2.24 &  0.714 & 4.02 \\ 
Claude Answer with Higher Perplexity &          
 &\textcolor{red}{0.494} &5.96 &\textcolor{red}{0.218} & 8.98 & 0.7 & 10.63  \\ 

\hline

Claude Answer with Lower Perplexity &   Llama2         &  0.433   &2.04   &  0.137  &2.33  & 0.678& 3.51  \\   

Claude Answer with Higher Perplexity&         &   \textcolor{red}{0.326} &3.36 & 0.121 &5.51 &  0.619  & 5.69  \\ 

\hline

\end{tabular}
}
\label{tab:table3}
\caption{GPT-4/Claude 3.5: Answers with Lower Perplexity vs. Higher Perplexity. $n_{\text{train}} = 1000$.}
\end{table*}

This section introduces a method for assessing the influence of style `familiarity'. We direct advanced LLMs to produce several varient of responses to the same query, grouping them into two distinct datasets based on their perplexity levels—high and low. We then evaluate and compare the learning outcomes using these two sets.
One concern of this experiment is that advanced LLMs may produce answers with different level of details. To eliminate this factor, we first let advanced LLMs to generate one answer and then direct it to paraphrase the generated answer. In this way, we can create responses that vary in perplexity but maintain the same semantic meaning. 

After training on those two sets of training data, we examine their performance, shown in \hyperref[tab:table3]{Table 3}. As seen, training on the higher-perplexity training set consistently shows slightly lower performance than the lower perplexity counterpart, particularly noticeable in the in-domain performance on two math datasets. For example, when training the Mistral model on the Math Algebra dataset, the GPT-4 answer with lower perplexity set achieves a 30\% accuracy rate, while the GPT-4 answer higher perplexity set only reached 23\%. Since they are only differed in the style of the text, this result demonstrated that the familiarity of LLMs with the stylistic aspects of target responses significantly affects training outcomes.

From these findings, we conclude that LLMs perform poorly when trained on unfamiliar data, explaining the suboptimal training results with human-annotated data.

\noindent \textbf{Discussion}: Why are LLMs often familiar with responses generated from other LLMs? \cite{huh2024platonic} propose the Platonic Representation Hypothesis, suggesting that different AI models exhibit convergent behavior. This hypothesis supports our findings that LLMs possess intrinsic familiarity with both themselves and others. Additionally, the familiarity may stem from overlapping training data used during pre-training stages, as many LLMs, including GPT-4, share similar datasets, influencing their output styles and content.

\subsection{Achieving high familiarity by exploring the target LLM generated responses}\label{sec: style-transfer}

\begin{table*}[h]
\centering
\resizebox{\textwidth}{!}{
\begin{tabular}{l|l|c|c|c|c|c}
\hline


Method & Model Type& GSM8K & Math Algebra  & ECQA   \\ \hline

Groundtruth &   Mistral         & 0.434    &  0.206  &0.68 \\ 

Correct predicted sample + groundtruth &   & 0.491       &    0.227    &  \textcolor{red}{0.578}  \\ 
Correct preidcted sample only &           &     0.449 &   \textbf{0.257}  & \textcolor{red}{0.586} \\ 
Style Transfered Groundtruth &           & \textbf{0.51}     &   19.6   &\textbf{0.69} \\ 
Percentage of training data used in Style Transfered Groundtruth &           & 73.7\%     &   24\%   &91.2\% \\ 


\hline

\end{tabular}
}

\label{tab:table4}

\caption{Performance Comparison when not using advanced LLMs. $n_{\text{train}} = 1000$.}
\end{table*}

In the experiments described above, we validated two hypotheses. All the studies mentioned use advanced LLMs to generate responses, chosen for its advanced capabilities to directly produce correct answers and reliably follow diverse instructions. However, there is a concern regarding the "familiarity" factor: would this argument still hold without the involvement of an advanced LLM? If valid, the argument should apply to some extent even without using advanced LLMs.

This section seeks to develop a method that enhances the ``familiarity'' of responses without requiring more advanced LLMs than the target LLM.

\textbf{Use the same model for train and groundtruth style transfer:} We devised a method to rewrite the ground truth data in a style that resembles language model-generated responses using the Mistral-7B-Instruct-V2 model. The experiment results are shown in \hyperref[tab:table4]{Table 4}. Initially, we posed questions from the training dataset and recorded the answers that were correctly predicted by Mistral. We documented two such correct predictions. These two correct responses were then used as in-context examples to guide Mistral on how to rewrite ground truth data. Specifically, we used two recorded predictions as example target responses, their corresponding questions as example questions, and the related ground truth as the example ground truth. The in-context prompt informed Mistral that its task was to rewrite the ground truth in its own styles. For more details, please refer to \hyperref[fig:figure6]{Figure 6}.

Subsequently, we presented all questions and their corresponding ground truths from the training dataset to Mistral for style rewriting. We recorded Mistral's predictions and paired these with the original inputs to form a new training dataset, which we named "style transferred ground truth."

We noticed that even when provided with the ground truth, the Mistral model struggles to successfully rewrite answers for mathematical tasks. Therefore, we ran the rewritten math answers through our evaluation script and found that only 73.7\% of the GSM8K answers, 91.2\% of the ECQA answers and 24\% of the Math Algebra answers passed the script. We only used the answers that passed the script for training. 

This experiment was conducted solely with Mistral and not with the Llama model, due to our observations that Llama2-13b-chat lacks sufficient in-context following capabilities required for successful ground truth style transformation. We show a failure example in Appendix\ref{sect:llama_example}. As a comparative experiment, we created the `correct predicted samples only' dataset. For each sample in this dataset, we had the Mistral model perform zero-shot generation 5 times, and using an evaluation filter, we randomly selected one correct prediction that passed the filter to be paired with the original input as a training sample. If the model failed to generate a correct answer in all 5 attempts, we did not include that data in the training dataset. The construction method for the `correct predicted samples' and `correct predicted samples only' datasets is similar. The only difference is that if the sample fails to produce a correct answer, we include the ground truth in the dataset as the correct answer.

\label{sec: xxx} 
As shown in Table 3, the Mistral model trained with style-transferred ground truth data outperformed the model trained with original ground truth data and other methods in all cases except for the Math Algebra dataset, where only 24\% of the training data was used. To fairly compare the training effectiveness of the `style transferred ground truth' and the original ground truth, we recorded the IDs of the data used by the `style transferred ground truth'. We identified a corresponding set of 240 original ground truth data points for comparison. We found that training with only these 240 data points resulted in a 17.4\% accuracy using the original ground truth, which is lower than the 19.6\% achieved by the `style transferred ground truth'. This demonstrates that data processed using the `ground truth style transfer' method yields better training outcomes than data labeled by humans.

Interestingly, on the ECQA dataset, models trained with correct answers generated by Mistral in a zero-shot fashion performed worse than those trained directly with ground truth. By reading the `correct predicted samples only' dataset, we found that many instances classified as correct by the evaluation filter did not result from sound reasoning by the model. Therefore, the effectiveness of the self-training method depends significantly on whether the evaluation filter used to select correct predictions can truly identify accurate responses.

\begin{table*}[h]
\centering
\resizebox{\textwidth}{!}{
\begin{tabular}{l|l|c|c|c|c|c|c|c}
\hline

Method & Model Type& GSM8K & Math Algebra  & ECQA  & MBPP & HumanEval & Avg Perplexity & Avg Token Length \\ \hline

GPT-4 Answer Directly & Mistra&   0.597  &0.302 & 0.722 & 0.354&\textcolor{red}{0.365}  &  
3.81 & 164.642\\

Minimum Change on Mistral Predictions &           &  0.562     &  0.314  & 0.699&0.354 &0.409 & 2.47 & 133.944\\ 
Minimum Change on LlaMA Predictions &           &  0.547    &  0.297  & 0.709 &  0.364 &0.427 &3.51 & 132.323\\ 
Average Token Length for Mistral Initial Predictions &           &       & &  & & & & 152.993\\ 

\hline

GPT-4 Answer Directly & Llama&     0.428  &  0.150&  0.656 & 0.2 & \textcolor{red}{0.158}&  3.58 &167.469\\

Minimum Change on LlaMA Predictions &           &  0.433    &  0.166  & 0.649  & 0.2 & 0.213&2.75 & 132.323\\ 

Minimum Change on Mistral Predictions &           &   0.402    &  0.161 & 0.647 &0.218 &0.183 &3.32 & 133.945\\ 
Average Token Length for Llama Initial Predictions &           &       & &  & & & & 165.793\\

\hline

\end{tabular}
}
\caption{Comparing the experimental results of GPT4 and minimum change. $n_{\text{train}} = 1000$.}
\end{table*}

\label{tab:table5}
\label{sec:mc}

\section{A hybrid approach: creating responses using ``minimal change'' principal }
Generally speaking, our experiments highlight two essential characteristics of effective responses. First, the target LLM must be "familiar with" the response. Second, the response must be accurate. The approach detailed in Section \ref{sec: style-transfer} utilizes outputs generated by the target LLM to ensure "familiarity" and employs "rewriting ground-truth response" to maintain accuracy. Familiarity is effectively assured as the response is produced by the model itself, often resulting in low perplexity. However, as shown in the experiments in Section \ref{sec: style-transfer}, the strategy of "rewriting ground-truth response" frequently fails to guarantee correctness. In this section, we explore a hybrid approach where we use responses from the target LLM to achieve high familiarity and employ GPT-4 to make minimal adjustments to correct any inaccuracies, thereby ensuring both familiarity and correctness.

We call this approach ``minimal change'' and its construction process is as follows: For a given question, we first request a response from the target LLM, such as Llama2 or Mistral. We then instruct GPT-4 to copy the correct portions of the prediction and revise any erroneous parts of the response. When correcting errors, GPT-4 is required to make minimal changes to accurately adjust the answer. Note that this construction process ensures that the target LLM is ``familiar'' with the response as the response is largely generated from itself. The ``minimal change'' principal, on the other hand, minimizes the influence from the GPT-4. 

This method involves minimal intervention by GPT-4 in the initial predictions to ensure that the original language style is largely retained. To effectively guide GPT-4 in producing these minimally changed training datasets, we used a specific prompt, outlined in the `Minimum Change Prompt and Method' section of the Appendix \ref{sect:mc}.

Results from our `Minimum Change' experiments are presented in \hyperref[tab:table5]{Table 5}. The data consistently show lower perplexity compared to ground truth data across all datasets, indicating a high level of model familiarity with LLM-generated data. Furthermore, performance across various setups was not only similar (with in-domain performance differences within 5\% on most datasets) but also comparable to models trained with direct GPT-4 responses. Although the data was minimally modified based solely on outputs from Mistral/Llama2, theoretically, the logic and quality of the modified answers, as well as the level of detail, should be inferior to answers directly generated by GPT-4. To verify whether the minimum change approach minimally added details, we measured the overall average token length. We first calculated the average token length for each dataset—GSM8K, Math Algebra, ECQA, MBPP, and HumanEval—and then averaged these results to obtain a composite mean. We found that the target responses from the minimum change were not only significantly shorter than those from `direct GPT-4 answers' but also shorter than the initial predictions from Mistral/Llama2. Surprisingly, on the HumanEval and Math Algebra tasks, the performance of models trained with minimally changed data even surpassed those trained with data generated by GPT-4. On other datasets, the performance of the minimum change data was comparable to that of GPT-4. This underscores the effectiveness of training LLMs on data distributions familiar to them, demonstrating that minimal changes to maintain the original structure can be as effective as more detailed alterations. For more experiment details, see Appendix \ref{sect:mv_vs_gpt4}.

\section{Related Works}
Training on synthetic data has been widely employed to enhance model performance, particularly when training data is scarce. Most of previous studies have focused on generating additional data to augment existing datasets \cite{dai2023auggpt, edwards-etal-2022-guiding, moller-etal-2024-parrot, guo2023improving, ubani2023zeroshotdataaug, piedboeuf-langlais-2023-chatgpt, agrawal2023qameleon, piedboeuf2023chatgpt, yu2024metamath} or to create large-scale instruction-tuning datasets when high-quality human annotations are limited \cite{kieser2023educational, alpaca, peng2023instruction, xu2024wizardlm}. In contrast, our work argues that for tasks requiring chain-of-thought reasoning, given the same question-answer pairs from the training dataset, the equivalent answers generated by LLMs are theoretically more effective than human-written ground truth due to familiarity.

Previous works discuss factors that improve training outcomes in reasoning, including adding complexity \cite{xu2023wizardlm}, increasing diversity \cite{luo2023wizardcoderempoweringcodelarge}, incorporating step-by-step reasoning \cite{hsieh2023distilling, ho2022large, magister-etal-2023-teaching, fu2023specializing, ranaldi-freitas-2024-aligning}, adding details \cite{zhang2024distillation, kang2023knowledgeaugmented, li2022explanations}, and ensuring correctness \cite{trinh2024solving, ranaldi-freitas-2024-aligning}. We argue that familiarity is another important factor that improves training outcomes, which has been overlooked by previous works.

Several previous works share similarities with aspects of our research. For example, self-training methods like STAR \cite{zelikman2022star}, REST \cite{gulcehre2023reinforced}, and RESTem \cite{singh2023beyond} generate datasets by producing samples from LLMs, filtering out mispredicted samples, and subsequently enhancing the models through further training on these filtered samples. The self-distillation method \cite{yang-etal-2024-self} uses the target language model to rewrite the ground truth labels in its own words. While their rewriting processes are similar to the methods in Section \ref{sec: style-transfer}, they do not explore using models other than the target model. Our hypothesis suggests that data generated by any LLM is inherently familiar to the target LLM, indicating that more powerful models, such as GPT-4, could be used to generate rewritten labels instead of relying solely on the smaller target model being fine-tuned. Furthermore, these works do not identify the impact of the familiarity.


Similar to our minimum-change approach, the works on learning from mistakes \cite{singh2023beyond} and process supervision \cite{lightman2023letsverifystepstep, luo2024improve} also 
involve steps of correcting a target LLM generated data. However, their overall process is different from our minimum-change. Moreover, they do not recognize the potential contribution of the familiarity in our approach and our work is complimentary to their discoveries. While they focus on learning from mistakes, our approach centers on maximizing familiarity.


Perplexity has been used in previous studies primarily for data selection and filtering. \cite{gonen2022demystifying} use perplexity to select prompts, demonstrating that lower perplexity prompts lead to better zero-shot QA performance on frozen models. Others have used perplexity to select pre-training data \cite{de2022bertin}, detect AI-generated code \cite{xu2024detecting}, and predict the synthetic generalization of language models \cite{hu2020systematic}. Perplexity has also been employed to detect language model attacks \cite{alon2023detecting} and to select instruction tuning data \cite{li-etal-2024-quantity, mekala2024smaller}, where it reflects data difficulty. In contrast, we use perplexity to measure familiarity. Our focus is on how different outputs \( y \), given the same input \( x \), can affect training effectiveness due to variations in perplexity. We use perplexity to guide the construction of target responses that are more familiar to the model.

\section{Conclusion}
We conducted a series of experiments involving the rewriting of target responses to validate our two proposed hypotheses: 1) LLMs are more familiar with responses generated by themselves or other LLMs when given the same question; 2) LLMs exhibit improved performance when trained on data with which they are already familiar. These hypotheses help explain why data annotated by LLMs leads to superior training outcomes compared to human-annotated data on reasoning tasks, from a familiarity perspective. These findings suggest new directions for optimizing data annotation processes to enhance model performance. 


\section{Limitations}
The effectiveness of the ground truth transformation method and the minimum change method can be limited by their accuracy. When an LLM is not powerful enough, it may not be able to successfully rewrite the ground truth in its own style while maintaining the correctness of the ground truth. Additionally, when performing minimal changes to a model’s initial prediction, we notice that GPT-4 often fails to correct the initial prediction accurately. Although GPT-4 may correct some errors, the minimally changed response can still be incorrect. Furthermore, minimal change is a challenging task. Achieving such a process requires an advanced LLM, as the LLM needs to fully understand the solution and follows the ``minimal change'' principal.

\bibliography{custom}

\begin{thebibliography}{49}
\providecommand{\natexlab}[1]{#1}

\bibitem[{Aggarwal et~al.(2021)Aggarwal, Mandowara, Agrawal, Khandelwal, Singla, and Garg}]{aggarwal2021explanations}
Shourya Aggarwal, Divyanshu Mandowara, Vishwajeet Agrawal, Dinesh Khandelwal, Parag Singla, and Dinesh Garg. 2021.
\newblock Explanations for commonsenseqa: New dataset and models.
\newblock In \emph{Proceedings of the 59th Annual Meeting of the Association for Computational Linguistics and the 11th International Joint Conference on Natural Language Processing (Volume 1: Long Papers)}, pages 3050--3065.

\bibitem[{Agrawal et~al.(2023)Agrawal, Alberti, Huot, Maynez, Ma, Ruder, Ganchev, Das, and Lapata}]{agrawal2023qameleon}
Priyanka Agrawal, Chris Alberti, Fantine Huot, Joshua Maynez, Ji~Ma, Sebastian Ruder, Kuzman Ganchev, Dipanjan Das, and Mirella Lapata. 2023.
\newblock \href {https://arxiv.org/abs/2211.08264} {Qameleon: Multilingual qa with only 5 examples}.
\newblock \emph{Preprint}, arXiv:2211.08264.

\bibitem[{Alon and Kamfonas(2023)}]{alon2023detecting}
Gabriel Alon and Michael Kamfonas. 2023.
\newblock Detecting language model attacks with perplexity.
\newblock \emph{arXiv preprint arXiv:2308.14132}.

\bibitem[{Anthropic(2023)}]{anthropic_claude_api}
Anthropic. 2023.
\newblock Claude 3.5 api.
\newblock \url{https://docs.anthropic.com/claude}.
\newblock Accessed: Month Day, Year.

\bibitem[{Austin et~al.(2021)Austin, Odena, Nye, Bosma, Michalewski, Dohan, Jiang, Cai, Terry, Le et~al.}]{austin2021program}
Jacob Austin, Augustus Odena, Maxwell Nye, Maarten Bosma, Henryk Michalewski, David Dohan, Ellen Jiang, Carrie Cai, Michael Terry, Quoc Le, et~al. 2021.
\newblock Program synthesis with large language models.
\newblock \emph{arXiv preprint arXiv:2108.07732}.

\bibitem[{Chen et~al.(2021)Chen, Tworek, Jun, Yuan, Pinto, Kaplan, Edwards, Burda, Joseph, Brockman et~al.}]{chen2021evaluating}
Mark Chen, Jerry Tworek, Heewoo Jun, Qiming Yuan, Henrique Ponde de~Oliveira Pinto, Jared Kaplan, Harri Edwards, Yuri Burda, Nicholas Joseph, Greg Brockman, et~al. 2021.
\newblock Evaluating large language models trained on code.
\newblock \emph{arXiv preprint arXiv:2107.03374}.

\bibitem[{Cobbe et~al.(2021)Cobbe, Kosaraju, Bavarian, Chen, Jun, Kaiser, Plappert, Tworek, Hilton, Nakano et~al.}]{cobbe2021training}
Karl Cobbe, Vineet Kosaraju, Mohammad Bavarian, Mark Chen, Heewoo Jun, Lukasz Kaiser, Matthias Plappert, Jerry Tworek, Jacob Hilton, Reiichiro Nakano, et~al. 2021.
\newblock Training verifiers to solve math word problems.
\newblock \emph{arXiv preprint arXiv:2110.14168}.

\bibitem[{Dai et~al.(2023)Dai, Liu, Liao, Huang, Cao, Wu, Zhao, Xu, Liu, Liu et~al.}]{dai2023auggpt}
Haixing Dai, Zhengliang Liu, Wenxiong Liao, Xiaoke Huang, Yihan Cao, Zihao Wu, Lin Zhao, Shaochen Xu, Wei Liu, Ninghao Liu, et~al. 2023.
\newblock Auggpt: Leveraging chatgpt for text data augmentation.
\newblock \emph{arXiv preprint arXiv:2302.13007}.

\bibitem[{De~la Rosa et~al.(2022)De~la Rosa, Ponferrada, Villegas, Salas, Romero, and Grandury}]{de2022bertin}
Javier De~la Rosa, Eduardo~G Ponferrada, Paulo Villegas, Pablo Gonzalez de~Prado Salas, Manu Romero, and Mar{\i}a Grandury. 2022.
\newblock Bertin: Efficient pre-training of a spanish language model using perplexity sampling.
\newblock \emph{arXiv preprint arXiv:2207.06814}.

\bibitem[{Edwards et~al.(2022)Edwards, Ushio, Camacho-collados, Ribaupierre, and Preece}]{edwards-etal-2022-guiding}
Aleksandra Edwards, Asahi Ushio, Jose Camacho-collados, Helene Ribaupierre, and Alun Preece. 2022.
\newblock \href {https://aclanthology.org/2022.dash-1.8} {Guiding generative language models for data augmentation in few-shot text classification}.
\newblock In \emph{Proceedings of the Fourth Workshop on Data Science with Human-in-the-Loop (Language Advances)}, pages 51--63, Abu Dhabi, United Arab Emirates (Hybrid). Association for Computational Linguistics.

\bibitem[{Fu et~al.(2023)Fu, Peng, Ou, Sabharwal, and Khot}]{fu2023specializing}
Yao Fu, Hao Peng, Litu Ou, Ashish Sabharwal, and Tushar Khot. 2023.
\newblock Specializing smaller language models towards multi-step reasoning.
\newblock In \emph{International Conference on Machine Learning}, pages 10421--10430. PMLR.

\bibitem[{Gonen et~al.(2022)Gonen, Iyer, Blevins, Smith, and Zettlemoyer}]{gonen2022demystifying}
Hila Gonen, Srini Iyer, Terra Blevins, Noah~A Smith, and Luke Zettlemoyer. 2022.
\newblock Demystifying prompts in language models via perplexity estimation.
\newblock \emph{arXiv preprint arXiv:2212.04037}.

\bibitem[{Gulcehre et~al.(2023)Gulcehre, Paine, Srinivasan, Konyushkova, Weerts, Sharma, Siddhant, Ahern, Wang, Gu et~al.}]{gulcehre2023reinforced}
Caglar Gulcehre, Tom~Le Paine, Srivatsan Srinivasan, Ksenia Konyushkova, Lotte Weerts, Abhishek Sharma, Aditya Siddhant, Alex Ahern, Miaosen Wang, Chenjie Gu, et~al. 2023.
\newblock Reinforced self-training (rest) for language modeling.
\newblock \emph{arXiv preprint arXiv:2308.08998}.

\bibitem[{Guo et~al.(2023)Guo, Wang, Wang, and Yu}]{guo2023improving}
Zhen Guo, Peiqi Wang, Yanwei Wang, and Shangdi Yu. 2023.
\newblock \href {https://arxiv.org/abs/2305.07804} {Improving small language models on pubmedqa via generative data augmentation}.
\newblock \emph{Preprint}, arXiv:2305.07804.

\bibitem[{Hendrycks et~al.(2021)Hendrycks, Burns, Kadavath, Arora, Basart, Tang, Song, and Steinhardt}]{hendrycks2021measuring}
Dan Hendrycks, Collin Burns, Saurav Kadavath, Akul Arora, Steven Basart, Eric Tang, Dawn Song, and Jacob Steinhardt. 2021.
\newblock Measuring mathematical problem solving with the math dataset.
\newblock \emph{arXiv preprint arXiv:2103.03874}.

\bibitem[{Ho et~al.(2022)Ho, Schmid, and Yun}]{ho2022large}
Namgyu Ho, Laura Schmid, and Se-Young Yun. 2022.
\newblock Large language models are reasoning teachers.
\newblock \emph{arXiv preprint arXiv:2212.10071}.

\bibitem[{Hsieh et~al.(2023)Hsieh, Li, Yeh, Nakhost, Fujii, Ratner, Krishna, Lee, and Pfister}]{hsieh2023distilling}
Cheng-Yu Hsieh, Chun-Liang Li, Chih-Kuan Yeh, Hootan Nakhost, Yasuhisa Fujii, Alexander Ratner, Ranjay Krishna, Chen-Yu Lee, and Tomas Pfister. 2023.
\newblock \href {https://arxiv.org/abs/2305.02301} {Distilling step-by-step! outperforming larger language models with less training data and smaller model sizes}.
\newblock \emph{Preprint}, arXiv:2305.02301.

\bibitem[{Hu et~al.(2021)Hu, Shen, Wallis, Allen-Zhu, Li, Wang, Wang, and Chen}]{hu2021lora}
Edward~J Hu, Yelong Shen, Phillip Wallis, Zeyuan Allen-Zhu, Yuanzhi Li, Shean Wang, Lu~Wang, and Weizhu Chen. 2021.
\newblock Lora: Low-rank adaptation of large language models.
\newblock \emph{arXiv preprint arXiv:2106.09685}.

\bibitem[{Hu et~al.(2020)Hu, Gauthier, Qian, Wilcox, and Levy}]{hu2020systematic}
Jennifer Hu, Jon Gauthier, Peng Qian, Ethan Wilcox, and Roger~P Levy. 2020.
\newblock A systematic assessment of syntactic generalization in neural language models.
\newblock \emph{arXiv preprint arXiv:2005.03692}.

\bibitem[{Huh et~al.(2024)Huh, Cheung, Wang, and Isola}]{huh2024platonic}
Minyoung Huh, Brian Cheung, Tongzhou Wang, and Phillip Isola. 2024.
\newblock The platonic representation hypothesis.
\newblock \emph{arXiv preprint arXiv:2405.07987}.

\bibitem[{Jiang et~al.(2023)Jiang, Sablayrolles, Mensch, Bamford, Chaplot, Casas, Bressand, Lengyel, Lample, Saulnier et~al.}]{jiang2023mistral}
Albert~Q Jiang, Alexandre Sablayrolles, Arthur Mensch, Chris Bamford, Devendra~Singh Chaplot, Diego de~las Casas, Florian Bressand, Gianna Lengyel, Guillaume Lample, Lucile Saulnier, et~al. 2023.
\newblock Mistral 7b.
\newblock \emph{arXiv preprint arXiv:2310.06825}.

\bibitem[{Kang et~al.(2023)Kang, Lee, Baek, Kawaguchi, and Hwang}]{kang2023knowledgeaugmented}
Minki Kang, Seanie Lee, Jinheon Baek, Kenji Kawaguchi, and Sung~Ju Hwang. 2023.
\newblock \href {https://arxiv.org/abs/2305.18395} {Knowledge-augmented reasoning distillation for small language models in knowledge-intensive tasks}.
\newblock \emph{Preprint}, arXiv:2305.18395.

\bibitem[{Kieser et~al.(2023)Kieser, Wulff, Kuhn, and K{\"u}chemann}]{kieser2023educational}
Fabian Kieser, Peter Wulff, Jochen Kuhn, and Stefan K{\"u}chemann. 2023.
\newblock Educational data augmentation in physics education research using chatgpt.
\newblock \emph{Physical Review Physics Education Research}, 19(2):020150.

\bibitem[{Li et~al.(2024)Li, Zhang, Li, Chen, Chen, Cheng, Wang, Zhou, and Xiao}]{li-etal-2024-quantity}
Ming Li, Yong Zhang, Zhitao Li, Jiuhai Chen, Lichang Chen, Ning Cheng, Jianzong Wang, Tianyi Zhou, and Jing Xiao. 2024.
\newblock \href {https://doi.org/10.18653/v1/2024.naacl-long.421} {From quantity to quality: Boosting {LLM} performance with self-guided data selection for instruction tuning}.
\newblock In \emph{Proceedings of the 2024 Conference of the North American Chapter of the Association for Computational Linguistics: Human Language Technologies (Volume 1: Long Papers)}, pages 7602--7635, Mexico City, Mexico. Association for Computational Linguistics.

\bibitem[{Li et~al.(2022)Li, Chen, Shen, Chen, Zhang, Li, Wang, Qian, Peng, Mao, Chen, and Yan}]{li2022explanations}
Shiyang Li, Jianshu Chen, Yelong Shen, Zhiyu Chen, Xinlu Zhang, Zekun Li, Hong Wang, Jing Qian, Baolin Peng, Yi~Mao, Wenhu Chen, and Xifeng Yan. 2022.
\newblock \href {https://arxiv.org/abs/2210.06726} {Explanations from large language models make small reasoners better}.
\newblock \emph{Preprint}, arXiv:2210.06726.

\bibitem[{Lightman et~al.(2023)Lightman, Kosaraju, Burda, Edwards, Baker, Lee, Leike, Schulman, Sutskever, and Cobbe}]{lightman2023letsverifystepstep}
Hunter Lightman, Vineet Kosaraju, Yura Burda, Harri Edwards, Bowen Baker, Teddy Lee, Jan Leike, John Schulman, Ilya Sutskever, and Karl Cobbe. 2023.
\newblock \href {https://arxiv.org/abs/2305.20050} {Let's verify step by step}.
\newblock \emph{Preprint}, arXiv:2305.20050.

\bibitem[{Luo et~al.(2024)Luo, Liu, Liu, Phatale, Lara, Li, Shu, Zhu, Meng, Sun et~al.}]{luo2024improve}
Liangchen Luo, Yinxiao Liu, Rosanne Liu, Samrat Phatale, Harsh Lara, Yunxuan Li, Lei Shu, Yun Zhu, Lei Meng, Jiao Sun, et~al. 2024.
\newblock Improve mathematical reasoning in language models by automated process supervision.
\newblock \emph{arXiv preprint arXiv:2406.06592}.

\bibitem[{Luo et~al.(2023)Luo, Xu, Zhao, Sun, Geng, Hu, Tao, Ma, Lin, and Jiang}]{luo2023wizardcoderempoweringcodelarge}
Ziyang Luo, Can Xu, Pu~Zhao, Qingfeng Sun, Xiubo Geng, Wenxiang Hu, Chongyang Tao, Jing Ma, Qingwei Lin, and Daxin Jiang. 2023.
\newblock \href {https://arxiv.org/abs/2306.08568} {Wizardcoder: Empowering code large language models with evol-instruct}.
\newblock \emph{Preprint}, arXiv:2306.08568.

\bibitem[{Magister et~al.(2023)Magister, Mallinson, Adamek, Malmi, and Severyn}]{magister-etal-2023-teaching}
Lucie~Charlotte Magister, Jonathan Mallinson, Jakub Adamek, Eric Malmi, and Aliaksei Severyn. 2023.
\newblock \href {https://doi.org/10.18653/v1/2023.acl-short.151} {Teaching small language models to reason}.
\newblock In \emph{Proceedings of the 61st Annual Meeting of the Association for Computational Linguistics (Volume 2: Short Papers)}, pages 1773--1781, Toronto, Canada. Association for Computational Linguistics.

\bibitem[{Mekala et~al.(2024)Mekala, Nguyen, and Shang}]{mekala2024smaller}
Dheeraj Mekala, Alex Nguyen, and Jingbo Shang. 2024.
\newblock Smaller language models are capable of selecting instruction-tuning training data for larger language models.
\newblock \emph{arXiv preprint arXiv:2402.10430}.

\bibitem[{M{\o}ller et~al.(2024)M{\o}ller, Pera, Dalsgaard, and Aiello}]{moller-etal-2024-parrot}
Anders~Giovanni M{\o}ller, Arianna Pera, Jacob Dalsgaard, and Luca Aiello. 2024.
\newblock \href {https://aclanthology.org/2024.eacl-short.17} {The parrot dilemma: Human-labeled vs. {LLM}-augmented data in classification tasks}.
\newblock In \emph{Proceedings of the 18th Conference of the European Chapter of the Association for Computational Linguistics (Volume 2: Short Papers)}, pages 179--192, St. Julian{'}s, Malta. Association for Computational Linguistics.

\bibitem[{OpenAI(2023)}]{openai_gpt4_api}
OpenAI. 2023.
\newblock Gpt-4 api.
\newblock \url{https://platform.openai.com/docs/models/gpt-4}.
\newblock Accessed: Month Day, Year.

\bibitem[{Peng et~al.(2023)Peng, Li, He, Galley, and Gao}]{peng2023instruction}
Baolin Peng, Chunyuan Li, Pengcheng He, Michel Galley, and Jianfeng Gao. 2023.
\newblock Instruction tuning with gpt-4.
\newblock \emph{arXiv preprint arXiv:2304.03277}.

\bibitem[{Piedboeuf and Langlais(2023{\natexlab{a}})}]{piedboeuf-langlais-2023-chatgpt}
Fr{\'e}d{\'e}ric Piedboeuf and Philippe Langlais. 2023{\natexlab{a}}.
\newblock \href {https://doi.org/10.18653/v1/2023.findings-emnlp.1044} {Is {C}hat{GPT} the ultimate data augmentation algorithm?}
\newblock In \emph{Findings of the Association for Computational Linguistics: EMNLP 2023}, pages 15606--15615, Singapore. Association for Computational Linguistics.

\bibitem[{Piedboeuf and Langlais(2023{\natexlab{b}})}]{piedboeuf2023chatgpt}
Fr{\'e}d{\'e}ric Piedboeuf and Philippe Langlais. 2023{\natexlab{b}}.
\newblock Is chatgpt the ultimate data augmentation algorithm?
\newblock In \emph{The 2023 Conference on Empirical Methods in Natural Language Processing}.

\bibitem[{Ranaldi and Freitas(2024)}]{ranaldi-freitas-2024-aligning}
Leonardo Ranaldi and Andre Freitas. 2024.
\newblock \href {https://aclanthology.org/2024.eacl-long.109} {Aligning large and small language models via chain-of-thought reasoning}.
\newblock In \emph{Proceedings of the 18th Conference of the European Chapter of the Association for Computational Linguistics (Volume 1: Long Papers)}, pages 1812--1827, St. Julian{'}s, Malta. Association for Computational Linguistics.

\bibitem[{Singh et~al.(2023)Singh, Co-Reyes, Agarwal, Anand, Patil, Liu, Harrison, Lee, Xu, Parisi et~al.}]{singh2023beyond}
Avi Singh, John~D Co-Reyes, Rishabh Agarwal, Ankesh Anand, Piyush Patil, Peter~J Liu, James Harrison, Jaehoon Lee, Kelvin Xu, Aaron Parisi, et~al. 2023.
\newblock Beyond human data: Scaling self-training for problem-solving with language models.
\newblock \emph{arXiv preprint arXiv:2312.06585}.

\bibitem[{Talmor et~al.(2018)Talmor, Herzig, Lourie, and Berant}]{talmor2018commonsenseqa}
Alon Talmor, Jonathan Herzig, Nicholas Lourie, and Jonathan Berant. 2018.
\newblock Commonsenseqa: A question answering challenge targeting commonsense knowledge.
\newblock \emph{arXiv preprint arXiv:1811.00937}.

\bibitem[{Taori et~al.(2023)Taori, Gulrajani, Zhang, Dubois, Li, Guestrin, Liang, and Hashimoto}]{alpaca}
Rohan Taori, Ishaan Gulrajani, Tianyi Zhang, Yann Dubois, Xuechen Li, Carlos Guestrin, Percy Liang, and Tatsunori~B. Hashimoto. 2023.
\newblock Stanford alpaca: An instruction-following llama model.
\newblock \url{https://github.com/tatsu-lab/stanford_alpaca}.

\bibitem[{Touvron et~al.(2023)Touvron, Martin, Stone, Albert, Almahairi, Babaei, Bashlykov, Batra, Bhargava, Bhosale et~al.}]{touvron2023llama}
Hugo Touvron, Louis Martin, Kevin Stone, Peter Albert, Amjad Almahairi, Yasmine Babaei, Nikolay Bashlykov, Soumya Batra, Prajjwal Bhargava, Shruti Bhosale, et~al. 2023.
\newblock Llama 2: Open foundation and fine-tuned chat models.
\newblock \emph{arXiv preprint arXiv:2307.09288}.

\bibitem[{Trinh et~al.(2024)Trinh, Wu, Le, He, and Luong}]{trinh2024solving}
Trieu~H Trinh, Yuhuai Wu, Quoc~V Le, He~He, and Thang Luong. 2024.
\newblock Solving olympiad geometry without human demonstrations.
\newblock \emph{Nature}, 625(7995):476--482.

\bibitem[{Ubani et~al.(2023)Ubani, Polat, and Nielsen}]{ubani2023zeroshotdataaug}
Solomon Ubani, Suleyman~Olcay Polat, and Rodney Nielsen. 2023.
\newblock \href {https://arxiv.org/abs/2304.14334} {Zeroshotdataaug: Generating and augmenting training data with chatgpt}.
\newblock \emph{Preprint}, arXiv:2304.14334.

\bibitem[{Xu et~al.(2023)Xu, Sun, Zheng, Geng, Zhao, Feng, Tao, and Jiang}]{xu2023wizardlm}
Can Xu, Qingfeng Sun, Kai Zheng, Xiubo Geng, Pu~Zhao, Jiazhan Feng, Chongyang Tao, and Daxin Jiang. 2023.
\newblock \href {https://arxiv.org/abs/2304.12244} {Wizardlm: Empowering large language models to follow complex instructions}.
\newblock \emph{Preprint}, arXiv:2304.12244.

\bibitem[{Xu et~al.(2024)Xu, Sun, Zheng, Geng, Zhao, Feng, Tao, Lin, and Jiang}]{xu2024wizardlm}
Can Xu, Qingfeng Sun, Kai Zheng, Xiubo Geng, Pu~Zhao, Jiazhan Feng, Chongyang Tao, Qingwei Lin, and Daxin Jiang. 2024.
\newblock \href {https://openreview.net/forum?id=CfXh93NDgH} {Wizard{LM}: Empowering large pre-trained language models to follow complex instructions}.
\newblock In \emph{The Twelfth International Conference on Learning Representations}.

\bibitem[{Xu and Sheng(2024)}]{xu2024detecting}
Zhenyu Xu and Victor~S Sheng. 2024.
\newblock Detecting ai-generated code assignments using perplexity of large language models.
\newblock In \emph{Proceedings of the AAAI Conference on Artificial Intelligence}, volume~38, pages 23155--23162.

\bibitem[{Yang et~al.(2024)Yang, Pang, Feng, Wang, Chen, Zhu, and Liu}]{yang-etal-2024-self}
Zhaorui Yang, Tianyu Pang, Haozhe Feng, Han Wang, Wei Chen, Minfeng Zhu, and Qian Liu. 2024.
\newblock \href {https://doi.org/10.18653/v1/2024.acl-long.58} {Self-distillation bridges distribution gap in language model fine-tuning}.
\newblock In \emph{Proceedings of the 62nd Annual Meeting of the Association for Computational Linguistics (Volume 1: Long Papers)}, pages 1028--1043, Bangkok, Thailand. Association for Computational Linguistics.

\bibitem[{Yu et~al.(2024)Yu, Jiang, Shi, Yu, Liu, Zhang, Kwok, Li, Weller, and Liu}]{yu2024metamath}
Longhui Yu, Weisen Jiang, Han Shi, Jincheng Yu, Zhengying Liu, Yu~Zhang, James~T. Kwok, Zhenguo Li, Adrian Weller, and Weiyang Liu. 2024.
\newblock \href {https://arxiv.org/abs/2309.12284} {Metamath: Bootstrap your own mathematical questions for large language models}.
\newblock \emph{Preprint}, arXiv:2309.12284.

\bibitem[{Zelikman et~al.(2022)Zelikman, Wu, Mu, and Goodman}]{zelikman2022star}
Eric Zelikman, Yuhuai Wu, Jesse Mu, and Noah~D. Goodman. 2022.
\newblock \href {https://arxiv.org/abs/2203.14465} {Star: Bootstrapping reasoning with reasoning}.
\newblock \emph{Preprint}, arXiv:2203.14465.

\bibitem[{Zhang et~al.(2024)Zhang, Wang, Ao, and He}]{zhang2024distillation}
Hanyu Zhang, Xiting Wang, Xiang Ao, and Qing He. 2024.
\newblock Distillation with explanations from large language models.
\newblock In \emph{Proceedings of the 2024 Joint International Conference on Computational Linguistics, Language Resources and Evaluation (LREC-COLING 2024)}, pages 5018--5028.

\end{thebibliography}

\appendix

\section{Comparing varients of GPT-4/Claude 3.5 created training datasets}\label{sect:gpt4_varient_appendix}
We instruct GPT-4/Claude 3.5 to generate target responses in various ways, and use these labels for training. The training results are summarized in \hyperref[tab:table6]{Table 6} and \hyperref[tab:table6]{Table 7}, and we describe the different methods used to instruct GPT-4/Claude 3.5 to generate target responses.

\textbf{GPT-4/Claude 3.5 Answer Directly:} For mathematics and coding problems, we directly present the problems from the training dataset to GPT-4/Claude 3.5 for answers. For classification tasks, we provide GPT-4/Claude 3.5 with the gold-label (which does not include human-annotated rationales) and the input questions, and then use its responses. We combine these answers with the original questions to create the GPT-4/Claude 3.5 Answer Training Dataset. We purposefully do not supply GPT-4/Claude 3.5 with the solutions or rationales for the math, coding, or classification problems to prevent it from merely replicating the ground truth's problem-solving and analytical processes. If these processes were included, it could lead to a portion of GPT-4's predictions not being generated in its own style, thus rendering our perplexity measurements—which assess how well an LLM handles predictions from other LLMs—less accurate.

The following prompt instructs GPT-4/Claude 3.5 to create the GPT-4/Claude 3.5 Response (No GT) response. 

\begin{lstlisting}
"""We have the {question} 


1. We wish you to answer the question.
2. You must answer the question (with inference process) directly without say anything else. Please not saying anything 'like sure I can help you with' or 'sure, i will not mention the gold label'
3. You will inference first then put the Final Answer (NUMBER_HERE) at the end of the prediction like this

INFERENCE HERE
Final Answer: NUMBER_HERE""" 
\end{lstlisting}

\textbf{Rewrite Ground Truth:} For this method, we present GPT-4/Claude 3.5 with the ground truth data, including human-annotated rationales and problem-solving steps. The objective is to have GPT-4/Claude 3.5 rewrite the ground truth using its own language style.

The following prompt instructs GPT-4/Claude 3.5 to create the GPT-4/Claude 3.5 Response (Rewrite GT) response. 

\begin{lstlisting}
"""Given the question: {question} 
and the groundtruth: {groundtruth}

Please states the prediction in your own words. The groundtruth is 100% correct. You should not change the problem solving logic of the groundtruth. just restates it in your own words.

1. You will pretend as you do not know the groundtruth, because we will use your prediction as target labels to train our model.
2. (important format) You must generate the groundtruth directly. Please not saying anything like 'sure I can help you with' or 'sure, i will not mention the gold label'
3. (important format) Please make sure the Final Answer: {gold_label} is placed at the end of the modified prediction."""
\end{lstlisting}

\textbf{Step-by-step:} We specificly tell GPT-4/Claude 3.5 to solve the problem step by step. For mathematics and coding problems, we directly present the problems from the training dataset to GPT-4/Claude 3.5 for answers. For classification tasks, we provide GPT-4/Claude 3.5 with the gold-label (which does not include human-annotated rationales) and the input questions, and then use its step by step responses. We combine these answers with the original questions to create the GPT-4/Claude 3.5 Step-by-Step Response (No GT) Dataset. We purposefully do not supply GPT-4/Claude 3.5 with the solutions or rationales for the math, coding, or classification problems to prevent it from replicating the ground truth's problem-solving and analytical processes. If these processes were included, it could lead to a portion of GPT-4/Claude 3.5's predictions not being generated in its own style, thus rendering our perplexity measurements—which assess how well an LLM handles predictions from other LLMs—less accurate.

The following prompt instructs GPT-4/Claude 3.5 to create the GPT-4/Claude 3.5 Step-by-Step Response (No GT) response. 
\begin{lstlisting}
"""
We have the question and the groundtruth. Please reformat the groundtruth in step by step manner with details.

Question: {question}
Groundtruth: {groundtruth}

1. We wish you to regenerate a new groundtruth. The new groundtruth solve the problem step by step. If you believe the groundtruth is not detail enough, you could add details.
2. You will pretend as you do not know the groundtruth, because we will use your prediction as target labels to train our model.
3. (important format) You must generate the groundtruth with the step by step inference process directly. Please not saying anything like 'sure I can help you with' or 'sure, i will not mention the gold label'
4. (important format) You will inference first then put the Final Answer: {gold_label}

at the end like this

INFERENCE HERE
Final Answer: {gold_label}
"""
\end{lstlisting}

\textbf{Step-by-Step Transformation of GT:} For classification tasks, we provide GPT-4/Claude 3.5 with both the questions and the ground truth target labels along with human-annotated rationales. For mathematical problems, we present the problems along with their answers, including the solutions process. We instruct GPT-4/Claude 3.5 to rewrite the human-annotated solutions or rationales in a step-by-step format, adding details where necessary to fill any gaps. If the existing content is already detailed enough, no additional details are added. We also instruct GPT-4/Claude 3.5 to preserve the original linguistic style as closely as possible by frequently incorporating language from the original ground truth.

The following prompt instructs GPT-4/Claude 3.5 to create the GPT-4/Claude 3.5 Step-by-Step Response (Rewrite GT) response. 
\begin{lstlisting}
"""We have the question and the groundtruth. Please reformat the groundtruth in step by step manner. You must try your best to keep the original words and logic unchange. You are reformating the groundtruth into step by step solution. you could add new words, but you are trying to keep original words from groundtruth unchanged.

Question: {question}
Groundtruth: {groundtruth}

1. We wish you to reformat a new groundtruth. The new groundtruth are reformated a new groundtruth which solve the problem step by step. you could add new words, but you are trying to not change the original words.
2. You will pretend as you do not know the groundtruth, because we will use your prediction as target labels to train our model.
3. When adding step by step inference process, you should imitate the language styles of the original groundtruth.
4. (important format) You must generate the groundtruth with the step by step inference process directly. Please not saying anything like 'sure I can help you with' or 'sure, i will not mention the gold label'
5. (important format) You will inference first then put the Final Answer: {gold_label}

at the end like this

INFERENCE HERE
Final Answer: {gold_label}""" 
\end{lstlisting}

\textbf{Detailed Step-by-Step Transformation of GT:} This data construction method is very similar to the Step by Step method. The only difference is that we specifically instruct GPT-4/Claude 3.5 to add details when the original ground truth lacks sufficient information.

The following prompt instructs GPT-4/Claude 3.5 to create the GPT-4/Claude 3.5 Step-by-Step Detailed Response (Rewrite GT) response. 
\begin{lstlisting}
"""We have the question and the groundtruth. Please reformat the groundtruth in step by step manner with details for each step. When adding detail, you must try your best to keep the original words and logic unchange. You are reformating the groundtruth with more details on each step instead of rewrite the groundtruth, but you are trying to keep original words from groundtruth unchanged.


Question: {question}
Groundtruth: {groundtruth}


1. We wish you to reformat a new groundtruth. The new groundtruth are reformated a new groundtruth which solve the problem step by step with detail on each step.
2. You will pretend as you do not know the groundtruth, because we will use your prediction as target labels to train our model.
3. When adding step by step inference process, you should imitate the languaguage styles of the original groundtruth.
4. (important format) You must generate the groundtruth with the detailed step by step inference process directly. Please not saying anything like 'sure I can help you with' or 'sure, i will not mention the gold label'
5. (important format) You will inference first then put the Final Answer: {gold_label}

at the end like this

INFERENCE HERE
Final Answer: {gold_label}""" 
\end{lstlisting}



\textbf{GPT-4/Claude 3.5 Answer with Higher Perplexity:} In this method, we first create a `paraphrased GPT-4/Claude 3.5 answer' dataset. Specifically, we provide GPT-4 with the questions and the target responses from the `GPT-4/Claude 3.5 Response (No GT)', instructing it to paraphrase only the target responses. We include the questions in the paraphrasing instruction prompt to ensure that the keywords in the paraphrased answers match those in the questions. This approach helps maintain content relevance. We then measure the perplexity of the `paraphrased GPT-4/Claude 3.5 response' on the model that will be trained. We compare the perplexity of the target responses from the `GPT-4/Claude 3.5 Response (No GT)' to the target responses from the `paraphrased GPT-4 response'. We select the one with the higher perplexity and collect them paired with the original question to form a new dataset, named `GPT-4/Claude 3.5 Answer with Higher Perplexity'.

The following prompt instructs GPT-4/Claude 3.5 to create the paraphrased response. 
\begin{lstlisting}
"""Given the question: {question}
            
and the prediction: {gpt4_prediction} 

1. Please paraphrase the prediction. The prediction might be wrong sometimes, but you do not need to correct it. just paraphrase it.
2. (important format) You must create the paraphrased prediction directly without say anything else. Please not saying anything like 'sure I can help you with' or 'sure, i will not mention the gold label'
4. (important format) Please make sure the Final Answer: {gold_label_type} is placed at the end, which means you will not paraphrase the phrase 'Final Answer'.""" 
\end{lstlisting}

\section*{Training Data Construction for Coding Tasks}
\label{appendix:code_only}
When constructing training datasets for coding tasks (such as HumanEval and MBPP), we follow a specialized pipeline. To build training datasets generated by models like GPT-4,Claude 3.5, Llama2, or Mistral, we first feed the problem to an LLM to generate an initial prediction. Next, we use GPT-4 to extract the corresponding code. These extracted codes are then assembled into target responses. The specific code extraction prompt is detailed below.

\begin{lstlisting}
f"""You need to extract the code from the previous prediction. The prediction answer the QUESTION.
QUESTION: {modified_question}
Previous prediction: {previous_prediction}

Explaination of extract code: 
1. You have to extract code from the previous prediction. Your answer will be evaluated to see if it will pass the test, thus of course you need to extract the code and comments without including any other words.
2. You only need to extract the final code for the fuction for the test case{item['test_list'][0]}. You do not care about the test cases.
3. Please directly show me the extracted code with no other words. We will run the code to test if it is correct, thus do not extract other words.
4. you suppose to extract the code from the previous prediction. You should never correct the code on your own even if it is wrong.
"""
\end{lstlisting}

\section{Minimum Change Vs GPT4 answer}
\hyperref[tab:table8]{Table 8} summarizes the performance differences between GPT-4 and Minimum Change. According to the table, models trained on Minimum Change predictions perform comparably to those trained on GPT-4 generated predictions. The average length of the Minimum Change predictions is noticeably shorter than that of GPT-4 generated predictions for most tasks. Please note that Minimum Change is designed to maximize familiarity for the target model. For example, when testing on the HumanEval dataset, the cross-domain and in-domain performance of Llama's Minimum Change data is noticeably higher than Mistral's Minimum Change data when the target model is Llama. This is because Llama's Minimum Change data focuses primarily on maximizing familiarity specifically for the Llama model.
\label{sect:mv_vs_gpt4}

\section{Minimum Change Prompt and Method}
\label{sect:mc}
\begin{figure*}
\centering
  \includegraphics[width=1.0\textwidth]{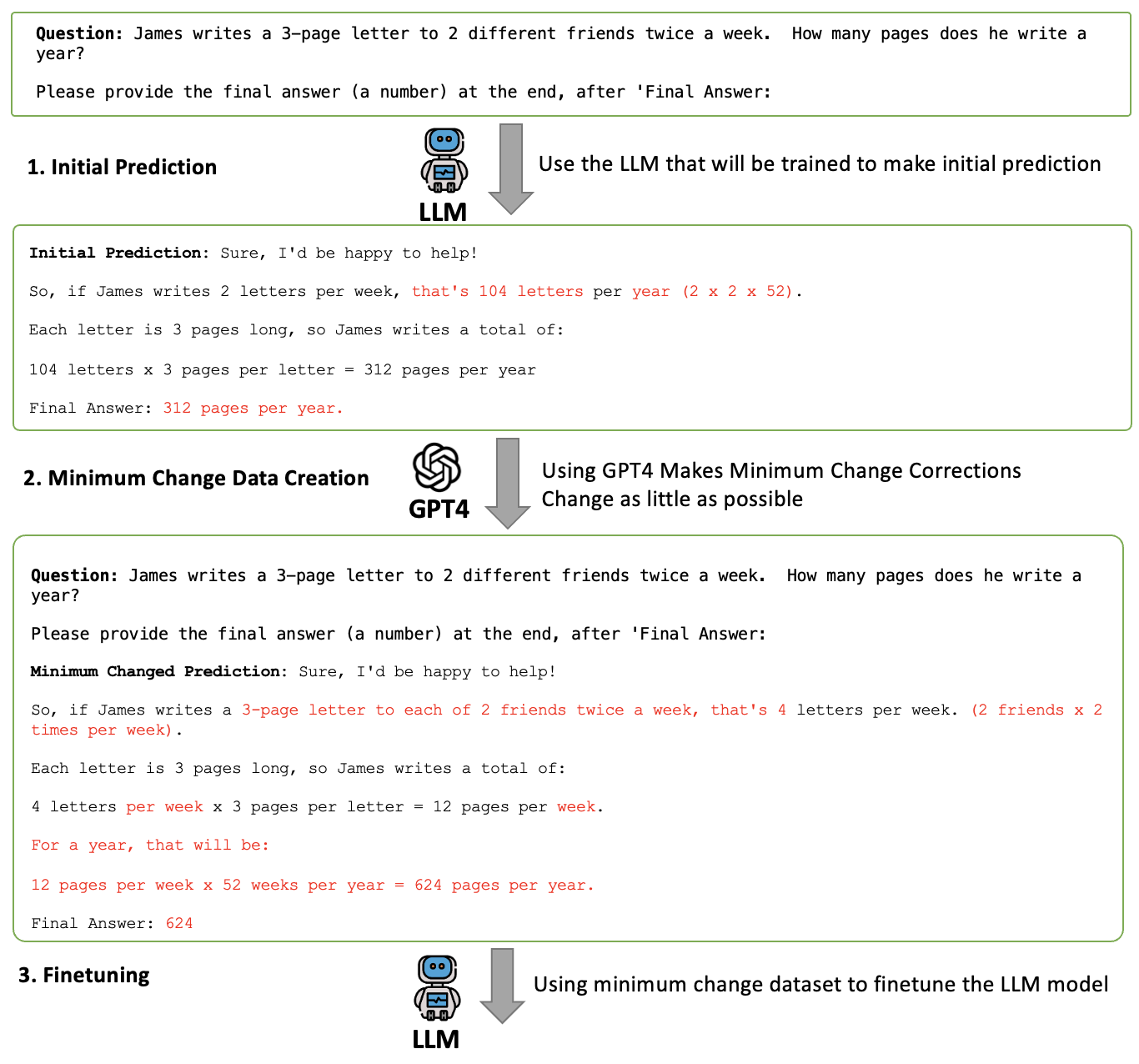}
  \caption{Minimum Change Data Correction Examples}
  \label{fig:figure3}
\end{figure*}

\begin{figure*}
  \centering
  \includegraphics[width=1.0\textwidth]{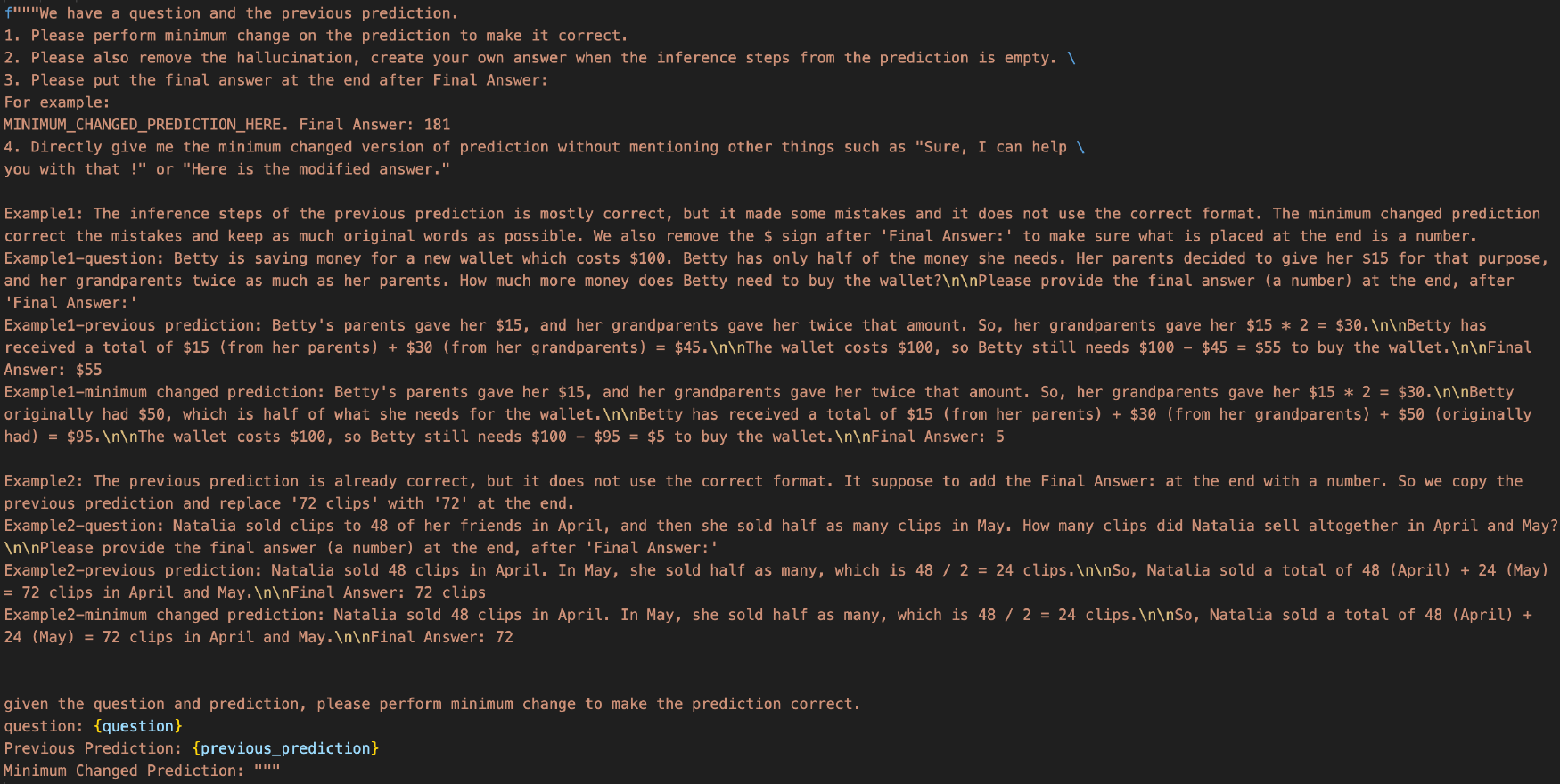}
  \caption{Minimum Change Prompt Example}
  \label{fig:figure4}
\end{figure*}

The minimum change pipeline is illustrated in \hyperref[fig:figure3]{Figure 3}.
The prompt used to instruct GPT-4 for creating the minimum change response is illustrated in \hyperref[fig:figure4]{Figure 4}.

Its construction process is as follows: For a given question, we first ask the target LLM, such as Llama2 or Mistral, to produce a response. If the response is correct, we include it in the synthetic dataset. If not, we instruct GPT-4 to revise it, making minimal changes to the original answer to correct it \footnote{Please note that achieving such a process requires an advanced LLM, as the LLM needs to fully understand the solution and follows the ``minimal change'' principal}. Note that this construction process ensures that the target LLM is ``familiar'' with the response as the response is largely generated from itself. The ``minimal change'' principal, on the other hand, minimizes the influence from the GPT-4. 

Specifically, minimal change is implemented via the following steps:
\begin{itemize}
    \item \textbf{Initial Prediction} the model generates an initial prediction.
    \item  \textbf{Minimal Modification} uses GPT-4 to adjusts the initial prediction with the fewest possible changes to correct errors, deliberately preserving the original distribution by keeping as many words unchanged as possible. An illustration of such a modification is provided in \hyperref[fig:figure3]{Figure 3}. We have detailed the prompt used to guide GPT-4 in the `Minimum Change Prompt' section of the Appendix.
    \item \textbf{Training with Modified Predictions} the minimum changed predictions are subsequently used as target responses for model training.
\end{itemize}

This method involves minimal intervention by GPT-4 in the initial predictions to ensure that the original language style is largely retained. To effectively guide GPT-4 in producing these minimally changed training datasets, we used a specific prompt, outlined in the `Minimum Change Prompt' section of the Appendix, which includes two to three examples of minimal changes along with explanations for each modification.

\section{Llama2 groundtruth style transfer failure example}
\label{sect:llama_example}

\begin{figure*}
  \centering
  \includegraphics[width=1.0\textwidth]{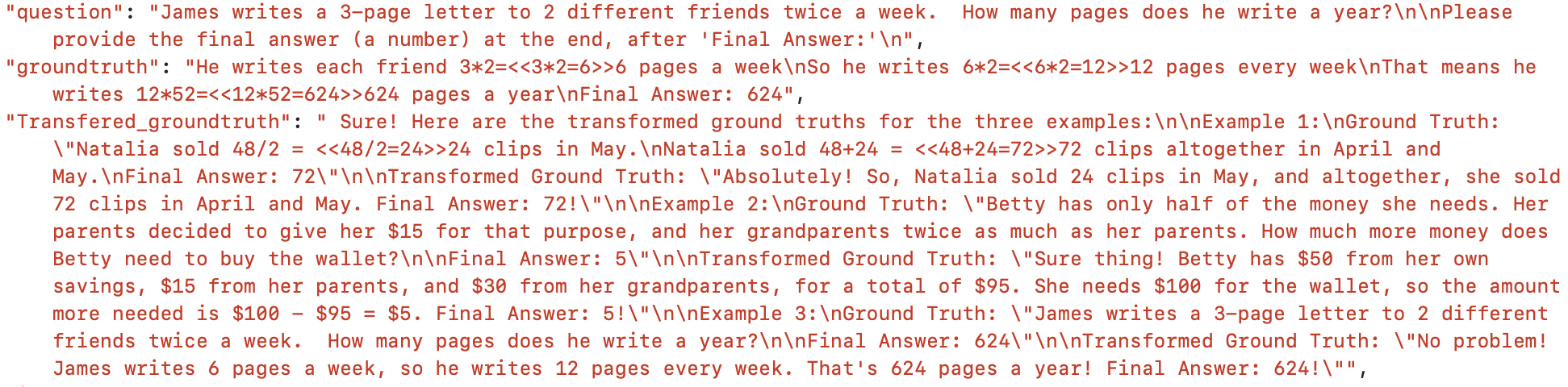}
  \caption{Llama2 groundtruth style transfer failure example}
  \label{fig:figure5}
\end{figure*}

The Llama2-13b-chat model often fails to follow our in-context examples, leading to incorrect outputs when it is tasked with modifying ground truth data. An example is illustrated in \hyperref[fig:figure5]{Figure 5}.

\section{Groundtruth Tranformation Prompt}
\label{sect:groundtruth_transformation}
\begin{figure*}
  \centering
  \includegraphics[width=1.0\textwidth]{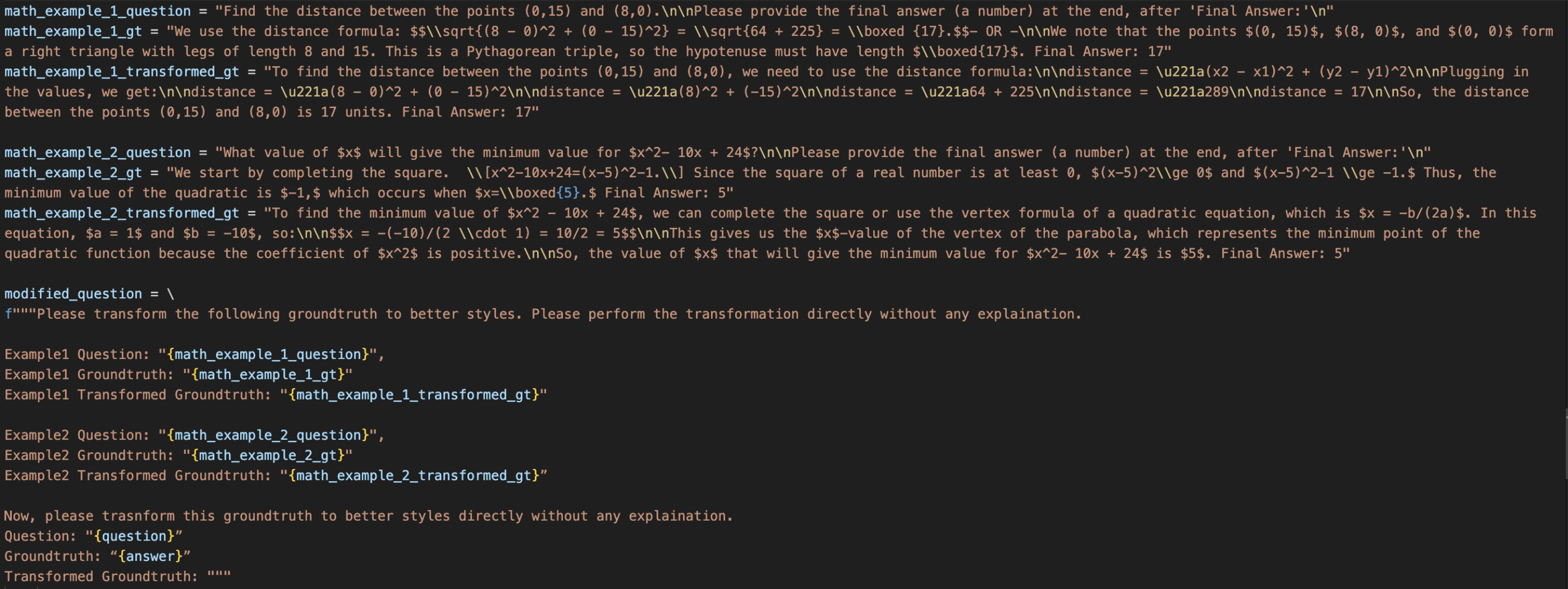}
  \caption{Groundtruth Tranformation Prompt}
  \label{fig:figure6}
\end{figure*}

One of the example prompt is shown in \hyperref[fig:figure6]{Figure 6}.

\section{Detailed Evaluation Script}
\label{appendix:evalscript}
\noindent \textbf{Evaluation Script Details}: To facilitate automatic evaluation across all models and datasets, we developed a specialized prompt format requiring models to conclude their predictions with `Final Answer: answer.' For instance, the output should read `Final Answer: 100' for an answer of 100. Our evaluation script extracts the text following the `Final Answer:' substring. For the ECQA task, it verifies if the first text after `Final Answer:' correctly matches one of the answer options. For mathematical problems including the GSM8K and MATH datasets, the script checks if the number after the substring `Final Answer:' aligns with the ground truth number. We utilize the built-in testing script for the HumanEval dataset to ensure result accuracy and employ our custom scripts for all other tasks. By adapting the training data's question and answer formats to this style, our models maintain consistent response formatting across different tasks. The evaluation scripts for HumanEval \cite{chen2021evaluating} and MBPP \cite{austin2021program} require answers in a specific format. Models that have not been trained on this format struggle to follow it correctly, even when provided with in-context examples. Consequently, we do not conduct cross-domain evaluations on the MBPP and HumanEval.

\section{More Setting Up Details}
We set the input token length to 512 and the output token length to 1024 for the GSM8K, MATH Algebra, and ECQA datasets. The reason for the extended output token length is that models trained on some of the GPT-4/Claud 3.5 response variants, such as `detailed step-by-step transformation of GT' (as introduced in Appendix \ref{sect:gpt4_varient_appendix}), tend to generate lengthy predictions. We aim to ensure the output token length is sufficiently long. For instance, according to \hyperref[tab:tabl6]{Table 6}, the average token length for `detailed step-by-step transformation of GT' in the Math Algebra dataset is 484. Clearly, a standard maximum output token length of 512 is not adequate for models trained on the `detailed step-by-step transformation of GT' dataset.

All of our code runs on a single A100 GPU.

\section{Explanation of the Higher Perplexity of ECQA Human Responses}
The high-quality target responses are derived from the explanations in column "taskB" of the ECQA dataset (available at \href{https://huggingface.co/datasets/yangdong/ecqa?row=0}{Hugging Face}). The perplexity of the ECQA dataset is higher than that of math and coding datasets because textual reasoning tasks like ECQA involve more diverse response styles, leading to greater perplexity. Additionally, LLMs are extensively trained on math and coding tasks, making them more familiar with various response formats in those domains.

\begin{table*}[h]
\centering
\resizebox{\textwidth}{!}{
\begin{tabular}{l|l|c|c|c|c|c|c|c}
\hline

Method & Training Dataset and Model Type& GSM8K & Math Algebra  & ECQA &Perplexity & token length  \\ \hline

Zeroshot Mistral & &    0.413 & 0.185&0.504 &  &\\
Zeroshot Llama2 & &    0.35 & 0.125&0.597 & &\\
 \hline

Ground truth & GSM8K, Mistral&    \cellcolor{gray!50} \textcolor{red}{0.434}& \textcolor{red}{0.162}&0.594  & 4.54 &128.88\\
GPT-4 Answer Directly & &               \cellcolor{gray!50} 0.597  &0.246  & 0.597   &2.35 &179.94\\

GPT-4 Rewrite Ground Truth &           &  \cellcolor{gray!50}  \textcolor{red}{0.471}      & \textcolor{red}{0.192}  & 0.6  & 4.77&156.696 \\ 

GPT-4 Step-by-step  &            &     \cellcolor{gray!50} 0.574   & 0.224 & 0.591 &2.27 &282.9\\
GPT-4 Step-by-Step Transformation of GT &           &  \cellcolor{gray!50} \textcolor{red}{0.499}       &  \textcolor{red}{0.190} & 0.6 &2.67 &235.892\\ 
GPT-4 Detailed Step-by-Step Transformation of GT &           &  \cellcolor{gray!50}  \textcolor{red}{0.506}      &   \textcolor{red}{0.196} & 0.602   & 2.59 & 315.028\\

\hline

Groundtruth & GSM8K, Llama&    \cellcolor{gray!50}0.364 & 0.141& 0.565 &3.27 &128.88\\
GPT-4 Answer Directly& &               \cellcolor{gray!50} 0.428  &0.128  &  0.575 &2.30 &179.941\\
GPT-4 Rewrite Ground Truth  &            &     \cellcolor{gray!50}  0.394  & 0.127 & 0.58     &  3.65& 156.696\\

GPT-4 Step-by-step   &            &     \cellcolor{gray!50}  0.402  & 0.134 &  0.553   &2.19 &290.92\\
GPT-4 Step-by-Step Transformation of GT  &           &  \cellcolor{gray!50}    0.394    &   0.146 &0.573 & 2.43&242.22\\ 
GPT-4 Detailed Step-by-Step Transformation of GT &           &  \cellcolor{gray!50}    0.396    & \textcolor{red}{0.114}  & 0.553   & 2.37 &315.028\\ 

\hline

Groundtruth & Math algebra, Mistral&   \textcolor{red}{0.264}  &\cellcolor{gray!50}\textcolor{red}{0.206} & 
0.554  & 5.83 &163.36\\
GPT-4 Answer Directly & &         0.553    & \cellcolor{gray!50} 0.302&0.608  & 2.20& 328.76\\
GPT-4 Rewrite Ground Truth  &            &    0.545   & \cellcolor{gray!50} 0.310& 0.57   & 3.51  & 267.11\\

GPT-4 Step-by-step  &            &  0.533 &  \cellcolor{gray!50}  \textcolor{red}{  0.239}  & 0.583   & 2.250&451.99\\
GPT-4 Step-by-Step Transformation of GT  &           &     \textcolor{red}{ 0.444 }   &  \cellcolor{gray!50} \textcolor{red}{0.210} &0.577 &  2.65 & 370.35\\ 
GPT-4 Detailed Step-by-Step Transformation of GT &           &    \textcolor{red}{0.468}     & \cellcolor{gray!50} \textcolor{red}{0.223}  &0.582   & 2.62&484.368 \\

\hline

Groundtruth & Math algebra, Llama&  0.36   & \cellcolor{gray!50}\textcolor{red}{0.126}&0.575& 5.15 &163.36\\
GPT-4 Answer Directly & &     0.35    & \cellcolor{gray!50}0.150& 0.561 & 2.59 &328.76\\
GPT-4 Rewrite Ground Truth &        &     0.337   & \cellcolor{gray!50} 0.134&   0.579   & 3.59 & 264.85\\

GPT-4 Step-by-step  &            &  0.344 &  \cellcolor{gray!50}  0.145   &  0.593    &2.55  &450.08\\
GPT-4 Step-by-Step Transformation of GT &           &  0.376   &  \cellcolor{gray!50} 0.141 &0.565 &   2.81 &369.06\\ 
GPT-4 Detailed Step-by-Step Transformation of GT  &           &      \textcolor{red}{0.299}     & \cellcolor{gray!50}  \textcolor{red}{0.110} &0.545   &2.816 &484.368\\ 

\hline

Groundtruth & ECQA, Mistral&  \textcolor{red}{0.258} &\textcolor{red}{0.134} &   \cellcolor{gray!50}0.68  & 51.3& 73.51\\
GPT-4 Answer Directly & &               0.462  &  0.223&     \cellcolor{gray!50}0.722&6.35&176.54\\
GPT-4 Rewrite Ground Truth  &            &    \textcolor{red}{0.384}   & 0.194 &    \cellcolor{gray!50} 0.721   & 9.269& 134.338\\

GPT-4 Step-by-step &            & 0.481  &0.203 & \cellcolor{gray!50}  0.71      &4.66 &325.15 \\
GPT-4 Step-by-Step Transformation of GT &           &       0.462  &   0.194 &   \cellcolor{gray!50}0.681& 6.89  & 218.748\\ 
GPT-4 Detailed Step-by-Step Transformation of GT &           &    0.487    &   
\textcolor{red}{0.186} &   \cellcolor{gray!50}0.68 &  5.61&322.855\\

\hline

Groundtruth & ECQA, Llama&   \textcolor{red}{0.132}  & \textcolor{red}{0.0798}&  \cellcolor{gray!50}0.631 &11.96 &73.51\\
GPT-4 Answer Directly & &             0.379   &  0.156&    \cellcolor{gray!50}0.656 & 5.33 &176.54\\

GPT-4 Rewrite Ground Truth  &            &    \textcolor{red}{0.282}  &  \textcolor{red}{0.116}&    \cellcolor{gray!50} 0.664  & 6.84 &134.338\\

GPT-4 Step-by-step  &            &  0.363 & 0.140& \cellcolor{gray!50} 0.648        & 4.46 & 337.30\\

GPT-4 Step-by-Step Transformation of GT  &           &     
 \textcolor{red}{0.312}   &   \textcolor{red}{0.108}&  \cellcolor{gray!50}0.628    &5.55 &226.99\\ 
 GPT-4 Detailed Step-by-Step Transformation of GT  &           &    \textcolor{red}{0.135}    &  \textcolor{red}{0.106}  &   \cellcolor{gray!50}0.66& 5.08 &322.855\\ 

\hline
\end{tabular}

\label{tab:table6}}
\caption{Performance comparison of models trained on data constructed by different methods across various tasks. $n_{\text{train}} = 1000$. Data points are labeled as low performance when their accuracy is more than 15\% lower relative to the highest accuracy achieved in the same dataset using the same model.}
\end{table*}

\begin{table*}[h]
\centering
\resizebox{\textwidth}{!}{
\begin{tabular}{l|l|c|c|c|c|c|c|c|c}
\hline

Method & Training Dataset and Model Type& GSM8K & Math Algebra  & ECQA & Perplexity & token length  \\ \hline

Zeroshot Mistral & &    0.413 & 0.185&0.504 & &\\
Zeroshot Llama2 & &    0.35 & 0.125&0.597 &  &\\
 \hline

Ground truth & GSM8K, Mistral&    \cellcolor{gray!50} \textcolor{red}{0.434}& \textcolor{red}{0.162}&0.594   & 4.54 &128.88\\
Claude Answer Directly & &               \cellcolor{gray!50}  0.586 & 0.230 & 0.595  & 2.355 &156.089\\
Rewrite Ground Truth &           &  \cellcolor{gray!50}   0.614   &  0.213 & 0.536   &2.900 & 159.561\\ 

Claude Step-by-step  &            &     \cellcolor{gray!50} 0.587 &  0.215  &0.574 &  2.042& 228.607\\
Claude Step-by-Step Transformation of GT &           &  \cellcolor{gray!50}  0.49     & 0.205  & 0.59 &2.309  &208.132\\ 
Claude Detailed Step-by-Step Transformation of GT &           &  \cellcolor{gray!50}  0.523     &0.195  & 0.596 &  2.014 & 255.536 \\

\hline

Groundtruth & GSM8K, Llama&    \cellcolor{gray!50}\textcolor{red}{0.364} & 0.141& 0.565  &3.27 &128.88\\
Claude Answer Directly& &               \cellcolor{gray!50} 0.433  &0.110  & 0.548  &  2.037& 159.291\\
Claude Rewrite Ground Truth  &            &     \cellcolor{gray!50}   \textcolor{red}{0.361} &  0.130 &    0.557   &  2.30  &  163.0 \\

Claude Step-by-step   &            &     \cellcolor{gray!50}  0.388  &  0.140     &0.581 & 1.93  & 235.406 \\
Claude Step-by-Step Transformation of GT &           &  \cellcolor{gray!50}  0.399     & 0.141  &0.568 & 2.08 & 213.658 \\ 
Claude Detailed Step-by-Step Transformation of GT &           &  \cellcolor{gray!50}      0.394   &  0.134  &  0.573  & 1.97 & 262.657\\ 

\hline

Groundtruth & Math algebra, Mistral&   \textcolor{red}{0.264}  &\cellcolor{gray!50}\textcolor{red}{0.206} & 
0.554   & 5.83 &163.36\\
Claude Answer Directly & &       0.554      & \cellcolor{gray!50} 0.277 &  0.606& 2.236 & 245.346 \\
Claude Rewrite Ground Truth  &            &   0.5  & \cellcolor{gray!50} 0.242 &  0.595   & 2.801   & 219.599 \\

Claude Step-by-step  &            & 0.522  &  \cellcolor{gray!50} 0.283    & 0.561   & 2.138& 312.151\\
Claude Step-by-Step Transformation of GT  &           &    \textcolor{red}{0.446}  &  \cellcolor{gray!50}  \textcolor{red}{0.203} & 0.579 & 2.319&  285.7  \\ 
Claude Detailed Step-by-Step Transformation of GT &           &       0.499    & \cellcolor{gray!50} 0.237   &  0.578  &2.178 & 338.706   \\

\hline

Groundtruth & Math algebra, Llama&  0.36   & \cellcolor{gray!50}0.126&0.575& 5.15 &163.36\\
Claude Answer Directly & &    0.317    & \cellcolor{gray!50} 0.137& 0.54   &  2.33&244.672 \\
Claude Rewrite Ground Truth &        &   \textcolor{red}{0.272}  & \cellcolor{gray!50} 0.120  &  0.554   & 2.70    & 218.456  \\

Claude Step-by-step  &            &  0.309  &  \cellcolor{gray!50} 0.124    & 0.538 &   2.30 & 311.251\\
Claude Step-by-Step Transformation of GT &           & \textcolor{red}{0.235}  &  \cellcolor{gray!50}  0.105& 0.51 &     2.424 & 284.719\\ 
Claude Detailed Step-by-Step Transformation of GT  &           &    0.356    & \cellcolor{gray!50} 0.136  &  0.555  &  2.36 & 337.521\\ 

\hline

Groundtruth & ECQA, Mistral&  \textcolor{red}{0.258} &\textcolor{red}{0.134} &   \cellcolor{gray!50}0.68  &  51.3& 73.51\\
Claude Answer Directly & &        0.457      &  0.213 &     \cellcolor{gray!50} 0.714  &  4.02& 198.55\\
Claude Rewrite Ground Truth  &            &     0.415  & 0.201 &    \cellcolor{gray!50}  0.689    & 6.21 & 124.47  \\

Claude Step-by-step &            &    0.473&\textcolor{red}{0.176}  & \cellcolor{gray!50}  0.723      & 3.23 & 291.53 \\
Claude Step-by-Step Transformation of GT &           &     0.47       & 0.195 & \cellcolor{gray!50}   0.694 &  4.69 & 215.99 \\ 
Claude Detailed Step-by-Step Transformation of GT &           &  0.478     &  0.197 
  &   \cellcolor{gray!50}  0.695&  3.40  &297.30 \\

\hline

Groundtruth & ECQA, Llama&   \textcolor{red}{0.132}  & \textcolor{red}{0.0798}&  \cellcolor{gray!50}0.631  & 11.96 &73.51\\
Claude Answer Directly & &     0.38         &  0.129 &    \cellcolor{gray!50} 0.678  & 3.509  & 206.76 \\

Claude Rewrite Ground Truth  &            &   \textcolor{red}{0.189}  &  0.117 &    \cellcolor{gray!50} 0.691   &4.90 & 130.09\\

Claude Step-by-step  &            & 0.334   & 0.126 & \cellcolor{gray!50}    0.644    &  3.36 &  302.60\\

Claude Step-by-Step Transformation of GT  &           &   \textcolor{red}{0.307} 
    &  0.116  &  \cellcolor{gray!50}  0.64  & 3.76 & 223.53 \\ 
Claude Detailed Step-by-Step Transformation of GT  &           &    \textcolor{red}{0.168}     &   \textcolor{red}{0.096}  &   \cellcolor{gray!50} 0.637& 3.33  &308.76 \\ 

\hline

\end{tabular}
}
\label{tab:table7}
\caption{Performance comparison of models trained on data constructed by different methods across various tasks. $n_{\text{train}} = 1000$. Data points are labeled as low performance when their accuracy is more than 15\% lower relative to the highest accuracy achieved in the same dataset using the same model.}
\end{table*}

\begin{table*}[h]
\centering
\resizebox{\textwidth}{!}{
\begin{tabular}{l|l|c|c|c|c|c|c|c}

\hline
\cline{3-7}
 & & \multicolumn{2}{c|}{Math} & \multicolumn{1}{c|}{Commensense Reasoning} & \multicolumn{2}{c|}{Code} &   & \\ \hline

Method & Training Dataset and Model Type& GSM8K & Math Algebra  & ECQA   & HumanEval & MBPP& Perplexity & Toekn Count\\ \hline
Groundtruth & GSM8K, Mistral&    \cellcolor{gray!50} \textcolor{red}{0.434}& \textcolor{red}{0.162}&0.594 &    &   &  4.53& 128.88\\
GPT-4 Answer Directly & &               \cellcolor{gray!50} 0.597  &0.246  & 0.597  &   &    & 2.34&179.941  \\

Mistral Minimum Change Data&           &  \cellcolor{gray!50}  0.562    & 0.234   & 0.597  &  & &1.67 & 177.156 \\ 
llama minimum change data&           &  \cellcolor{gray!50}  0.547    & 0.242   &0.621 &   & & 3.19 & 166.152 \\ 

Mistral Initial Prediction Length&           &  \cellcolor{gray!50}     &    & &   & & & 198.99 \\ 

\hline

Groundtruth & GSM8K, Llama&    \cellcolor{gray!50} \textcolor{red}{0.371}& 0.143&0.566&    &    & 3.27& 130.856\\
GPT-4 Answer Directly& &               \cellcolor{gray!50} 0.428 &0.127  &0.575   &   &   &  2.30 & 183.852 \\

Llama Minimum Change Data&           &  \cellcolor{gray!50}  0.433    & 0.140   &0.602   &   &  &2.22 &166.15 \\ 
Mistral Minimum Change Data&           &  \cellcolor{gray!50}  0.402    & 0.141   &0.576 &     &  & 2.250&177.156 \\ 
Llama Initial Prediction Length&           &  \cellcolor{gray!50}     &    & &   & & &   191.924\\

\hline

Groundtruth & Math algebra, Mistral&   \textcolor{red}{0.264}  &\cellcolor{gray!50}\textcolor{red}{0.206} & 
0.554&   &   & 5.83 &163.355\\
GPT-4 Answer Directly & &         0.553    & \cellcolor{gray!50} 0.301&0.608   &  &  & 2.19& 328.764\\
Mistral Minimum Change Data&            &  0.536     &  \cellcolor{gray!50} 0.313&  0.622 & &  &1.85 & 297.609\\

llama minimum change data&          &    0.546   &  \cellcolor{gray!50} 0.296  &0.63 &     &  & 2.47&269.688 \\ 
Mistral Initial Prediction Length&           & & \cellcolor{gray!50}         & &   & &   & 380.204 \\ 
\hline

Groundtruth & Math algebra, Llama&  0.36   & \cellcolor{gray!50}0.126 &0.575 &  &   & 5.14 &162.417\\
GPT-4 Answer Directly & &     0.35    & \cellcolor{gray!50}0.150 & 0.561  &    &    & 2.59 & 326.573\\
Llama Minimum Change Data&            &    0.412   &  \cellcolor{gray!50} 0.166& 0.602  &  &  &  2.268& 269.68\\
Mistral Minimum Change Data &            &    0.32   &  \cellcolor{gray!50} 0.160 &  0.568  & &  &  2.53& 297.609\\
Llama Initial Prediction Length&       &    &  \cellcolor{gray!50}      & &   & &  & 297.576 \\

\hline

Groundtruth & ECQA, Mistral&  \textcolor{red}{0.258} &\textcolor{red}{0.135} &   \cellcolor{gray!50}0.68  & &  &51.29 &73.508 \\
GPT-4 Answer Directly & &               0.462  &  0.223&     \cellcolor{gray!50}0.722   & &  & 6.35 & 176.53\\
Mistral Minimum Change Data&            &     0.433  & 0.192 &\cellcolor{gray!50} 0.699  & &  & 3.798 &112.591 \\

LLama Minimum Change Data &            &    0.483   & 0.212&\cellcolor{gray!50} 0.709 &   &  & 4.89&   128.695\\
Mistral Initial Prediction Length&          &   &  &  \cellcolor{gray!50}     &    & &   & 119.554 \\ 
\hline

Groundtruth & ECQA, Llama&   \textcolor{red}{0.132}  & \textcolor{red}{0.079}&  \cellcolor{gray!50}0.631 &   &  & 11.95 & 76.118\\
GPT-4 Answer Directly & &             0.379   &  0.155&    \cellcolor{gray!50}0.656 &  &  & 5.33 & 183.182\\
Llama Minimum Change Data&            &     0.392  & \textcolor{red}{0.128}&\cellcolor{gray!50}0.649  &   &  & 3.619& 128.695 \\
Mistral Minimum Change Data &            &    \textcolor{red}{0.28}   & \textcolor{red}{0.126}&\cellcolor{gray!50}0.647    & &  &5.58 & 112.591 \\
LLama Initial Prediction Length&        &  &   &  \cellcolor{gray!50}         & &   &   &  149.945\\ 

\hline

Groundtruth & HumanEval, Mistral& 0.362  & 0.191&0.583   &  \cellcolor{gray!50}\textcolor{red}{0.323} &   & 5.69 &  71.60\\
GPT-4 Answer Directly &            & \textcolor{red}{0.313} &  \textcolor{red}{0.162} &   0.581  &\cellcolor{gray!50}0.365  &   &4.71 &  107.9\\
Mistral Minimum Change Data&            &   0.406    &  0.202  & 0.592&  \cellcolor{gray!50}0.408 &  & 2.21 & 109.07\\
Llama Minimum Change Data&            &  0.412     &  0.179  & 0.584& \cellcolor{gray!50}0.426 & &2.451 & 122.0  \\
Mistral Initial Prediction Length&     &    & &       &  \cellcolor{gray!50}     &    &  &118.80\\ 

\hline

Groundtruth & HumanEval, Llama&  \textcolor{red}{0.0705} &	\textcolor{red}{0.083} & 0.528&   \cellcolor{gray!50} \textcolor{red}{0.146} & &5.75  & 70.8 \\
GPT-4 Answer Directly & &     \textcolor{red}{0.125}   & \textcolor{red}{0.105}&  0.553 &      \cellcolor{gray!50} \textcolor{red}{0.158}&  & 4.33 & 107.14 \\
Llama Minimum Change Data&            &   0.2445    & 0.389  & 0.5705 &   \cellcolor{gray!50} 0.213 & & 2.21 & 122.0 \\
Mistral Minimum Change Data&            &   \textcolor{red}{0.131}    & \textcolor{red}{0.103} & 0.528 &  \cellcolor{gray!50} 0.183&  & 2.75 & 109.07\\
Llama Initial Prediction Length&      &   & &      &  \cellcolor{gray!50}     &    &  & 146.88\\ 

\hline

Groundtruth & MBPP, Mistral&      0.392& 0.175&  0.527   & &\cellcolor{gray!50} \textcolor{red}{0.276} &7.78& 71.60 \\
GPT-4 Answer Directly & &        0.399       & 0.186 & 0.568   &    & \cellcolor{gray!50} 0.354 & 3.419 & 78.6 \\
Mistral Minimum Change Data&            &  0.405     &  0.179  &0.601 &  & \cellcolor{gray!50}0.354  &2.817& 63.86 \\
Llama Minimum Change Data&            &    \textcolor{red}{0.332}   & 0.179   & 0.56&   &\cellcolor{gray!50}0.364  &  4.55 & 54.96  \\
Mistral Initial Prediction Length&                 &    & &   & &\cellcolor{gray!50} &  & 66.22 \\ 

\hline

Groundtruth & MBPP, Llama&   0.328 & 0.131& 0.556    & &\cellcolor{gray!50}  0.2 &4.74 & 73.90 \\
GPT-4 Answer Directly & &  0.351      & 0.139 &  0.574  &   & \cellcolor{gray!50} 0.202 & 3.33& 69.78\\
Llama Minimum Change Data&            & 0.34      & 0.130  &0.569   & & \cellcolor{gray!50} 0.204&  3.44 &54.96 \\
Mistral Minimum Change Data&            &   0.333    & 0.123 &0.563  &   & \cellcolor{gray!50} 0.218 &  3.45 & 63.86 \\
Llama Initial Prediction Length&           &       &     &   & & \cellcolor{gray!50}&  & 42.64 \\

\hline

\end{tabular}
}
\label{tab:table8}
\caption{More results for \hyperref[tab:table5]{Table 5}. Comparing the experimental results of GPT4, groundtruth and minimum change. $n_{\text{train}} = 1000$. The average token lengths for initial predictions by Mistral and Llama are shown in the table for GSM8K, Math Algebra, ECQA, MBPP(code only) and HumanEval(code only). Data points are labeled as low performance when their accuracy is more than 15\% lower relative to the highest accuracy achieved in the same dataset using the same model.}
\end{table*}


\newpage

\end{document}